\documentclass[10pt,twocolumn,letterpaper]{article}

\usepackage[accsupp]{axessibility}

\PassOptionsToPackage{table}{xcolor}

\usepackage[pagenumbers]{iccv} 

\definecolor{gain}{rgb}{0.21,0.49,0.74}
\newcommand{\gain}[1]{\textcolor{gain}{#1}}
\newcommand{\res}[2]{{#1} {\small{({\gain{#2}})}}}
\usepackage{pifont}

\usepackage{multirow}
\usepackage{makecell}
\usepackage{adjustbox}
\usepackage{bbding}
\usepackage{pifont}
\usepackage{dsfont}
\definecolor{iccvblue}{rgb}{0.21,0.49,0.74}

\definecolor{aliceblue}{RGB}{240,248,255}
\DeclareMathOperator{\Softmax}{softmax}

\usepackage[pagebackref,breaklinks,colorlinks,allcolors=iccvblue]{hyperref}

\usepackage{tcolorbox}

\def\paperID{1044} 
\def\confName{ICCV}
\def\confYear{2025}

\title{Why LVLMs Are More Prone to Hallucinations in Longer Responses:\\The Role of Context}

{
\author{
Ge Zheng$^{1,2}$$^*$
\hspace{1.5em}
Jiaye Qian$^{2}$\thanks{Equal contribution.}
\hspace{1.5em}
Jiajin Tang$^{2}$
\hspace{1.5em}
Sibei Yang$^{1}$\thanks{Corresponding author is Sibei Yang.}
\\
{$^{1}$School of Computer Science and Engineering, Sun Yat-sen University}
\hspace{1.5em}
{$^{2}$ShanghaiTech University}
\\
Project Page: {\tt\small  \href{https://github.com/SooLab/HalTrapper}{https://github.com/SooLab/HalTrapper} }
}
}

\begin{document}
\maketitle

\begin{abstract}
Large Vision-Language Models (LVLMs) have made significant progress in recent years but are also prone to hallucination issues. They exhibit more hallucinations in longer, free-form responses, often attributed to accumulated uncertainties. In this paper, we ask: Does increased hallucination result solely from length-induced errors, or is there a deeper underlying mechanism? After a series of preliminary experiments and findings, we suggest that the risk of hallucinations is not caused by length itself but by the increased reliance on context for coherence and completeness in longer responses. Building on these insights, we propose a novel ``induce-detect-suppress" framework that actively induces hallucinations through deliberately designed contexts, leverages induced instances for early detection of high-risk cases, and ultimately suppresses potential object-level hallucinations during actual decoding. Our approach achieves consistent, significant improvements across all benchmarks, demonstrating its efficacy. The strong detection and improved hallucination mitigation not only validate our framework but, more importantly, re-validate our hypothesis on context. Rather than solely pursuing performance gains, this study aims to provide new insights and serves as a first step toward a deeper exploration of hallucinations in LVLMs' longer responses.
\end{abstract}

\section{Introduction}
\label{sec:intro}

Recently, Large Vision-Language Models (LVLMs)~\cite{bai2023qwen, chen2023shikra,chen2024internvl, liu2024improved, zhu2023minigpt, chen2023minigpt, instructblip} have made significant strides in developing general-purpose foundation models, achieving new, unprecedented capabilities. These models facilitate dynamic, context-driven interactions centered on the image content through open-ended conversations with users, given the input image and user instructions. Their impressive generative capabilities allow them to address various traditional vision tasks~\cite{radford2021learning, yuan2021florence, liu2023grounding, li2022grounded, li2022language, zou2023generalized, li2023blip, rombach2022high, 10638815, 9607461, zhu2025rethinking, shi2024part2object, 10005161, ge2018multievidencefilteringfusionmultilabel, Plain-Det, shi2024the} within a unified framework and seamlessly handle more comprehensive tasks~\cite{dai2025freeflyenhancingflexibility, dai2025adaptivelearningfinegrainedgeneralized, zheng2023ddcot, freebloom, Liu_Wen_Yang_2023_CCQ, yu2024seqafford, wu2023grounded, Spatial_and} that require world knowledge and complex reasoning, such as visual question answering~\cite{VQA, schwenk2022okvqa, hudson2019gqa, Shi_2023_LogoPrompt}, video-based reasoning~\cite{chen2023videollm, li2024mvbench, li2024llama, chen2024sharegpt4video} and mathematical reasoning~\cite{wang2024measuring, lu2023mathvista}. However, LVLMs also grapple with the hallucination issue~\cite{zhang2023language, ji2023survey, zhou2023analyzing, rohrbach2018object}, a serious and well-recognized challenge in deploying them in real-world scenarios~\cite{li2025cityanchor,  WildRefer, RealDex, He_Yang_Li_Li_Chang_Yu_2019}, due to their propensity for erroneous generation.

\begin{figure}
    \centering
    \vspace{-0.5mm}
    \includegraphics[width=0.99\linewidth]{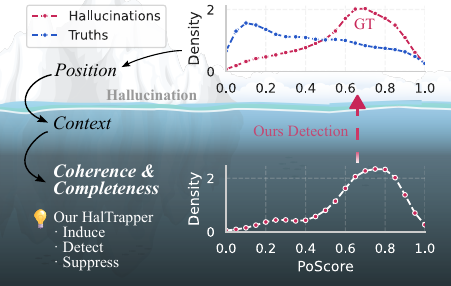}
    \vspace{-1mm}
    \caption{Left: Our three main findings and the three steps of our HalTrapper method. Right: The distribution of hallucination locations detected by our HalTrapper is close to the true distribution of hallucinations, indicating that our method, to some extent, captures the essence of LVLM hallucinations.}
    \label{fig:intro}
    \vspace{-4.5mm}
\end{figure}

Hallucination in LVLMs specifically refers to the discrepancy between the generated textual responses and the actual visual content and user instruction received, resulting in the production of irrelevant or non-existent objects, attributes, and other details. Various approaches have been proposed to reduce hallucinations, including filtering more reliable training data~\cite{liu2023mitigating, zhang2024reflective, zhou2023analyzing} or using specialized contrastive training materials~\cite{jiang2024hallucination} to re-fine-tune the model, thereby minimizing factually incorrect outputs. Rather than relying on costly, data-intensive solutions, recent approaches propose training-free strategies, such as contrastive decoding to contrastive model responses with their error-prone versions~\cite{leng2024mitigating, kim2024code, wang2024mitigating}, rolling back uncertain outputs~\cite{huang2024opera}, or enhancing attention to visual content~\cite{liu2024paying}. This has significantly mitigated the hallucination phenomenon, particularly in answering visual questions and identifying specific object hallucinations.
However, most of these efforts primarily focus on short responses, while hallucinations in long-form generation remains underexplored.

In this paper, we explore a seemingly straightforward—even widely taken for granted—phenomenon: LVLMs are more prone to hallucinations in longer, free-form textual responses compared to shorter answers. As shown in Fig.~\ref{fig:intro}, the frequency of hallucinated objects correlates with their position in the output token sequence, with a higher likelihood of appearing at later positions. Previous work~\cite{zhou2023analyzing} has also observed similar phenomena, simply attributing the issue to autoregressive text generation, where increasing length leads to accumulated hallucinations and greater uncertainties. However, beneath the intuitive manifestation of length (like an iceberg), deeper factors (beneath the surface) have yet to receive adequate attention: \textit{Is the increased hallucination merely a result of the cumulative errors due to length itself, or does it arise from a deeper underlying mechanism?}

Motivated by this, this paper presents the first and preliminary attempt to explore the underlying factors through a three-step analysis approach:
\begin{itemize}
    \item Phenomenon discovery to propose hypotheses (Sec.~\ref{sec:explore}).
    \item Preliminary statistics to analyze hypotheses (Sec.~\ref{sec:analysis}).
    \item Hypothesis application to detect and mitigate hallucinations, thereby re-validating it (Sec.~\ref{sec:method}).
\end{itemize}

\noindent\textbf{Phenomenon Discovery: Context may be a potential factor.}
Since free-form textual responses lack a predefined answer set or clear response forms, LVLMs rely heavily on context, including user instructions, visual input, and especially prior textual outputs. Consequently, we investigate the effect of context (see Sec.~\ref{sec:explore}), specifically by modifying either the image or text context and observing marked shifts in the distribution of the hallucination-length curve, which indicates that hallucinations appear at earlier positions.

\noindent\textbf{Hypothesis Analysis: Contextual coherence and completeness induce hallucinations.} Based on this observation, we hypothesize that contextual cues influence hallucinations along two key dimensions:
\begin{itemize}
    \item \textbf{Contextual coherence} drives LVLMs to maintain consistency with prior outputs while avoiding redundancy through distinct generation. The former focuses attention on contextual image content, while the latter shifts it to new information, potentially leading to dispersed attention, confusion, and hallucinations (see Sec.~\ref{sec:analysis_repetition}). Non-hallucinated tokens exhibit clear, focused attention, whereas hallucinated tokens show dispersed patterns. Notably, hallucinated tokens share highly similar attention distributions (see Fig.~\ref{fig:compare_attn}), suggesting LVLMs may be forced to attend to the same ungrounded, fragmented regions when balancing contextual and distinct content fails.
    \item \textbf{Contextual completeness} requires responses to incorporate comprehensive content while maintaining a logically coherent linguistic structure. However, when available recognized content is insufficient, LVLMs may employ contextual extrapolation as a compensatory strategy, potentially leading to hallucinated outputs (see Sec.~\ref{sec:analysis_expansion}). As contextual completeness increases, hallucinations tend to appear earlier in the response (see Fig.~\ref{fig:ee}). Furthermore, contextual extrapolation seems to follow inherently fixed patterns, with different sets of prompts repeatedly generating overlapping hallucinated tokens.
\end{itemize}

\noindent\textbf{Application and Re-validation.}
To further validate the hypotheses, we propose HalTrapper—a novel \textbf{``induce-detect-suppress"} framework that directly induces hallucinations by applying the two hypotheses, leverages the induced instances to detect high-risk cases \textbf{\textit{early to nip them in the bud}}, and ultimately suppress potential hallucinations during \textbf{\textit{the actual decoding stage.}}
\begin{itemize}
    \item \textbf{Induction}: (1) Imposing new, coherent outputs on an already complete response induces intra-response hallucinations. (2) Explicitly guiding imagination both based on and beyond recognized objects induces external expansion hallucinations.
    \item \textbf{Detection}: (1) Building on our coherence findings in Fig.~\ref{fig:compare_attn}, we identify hallucinations by analyzing attention similarity with induced intra-response hallucinations. (2) Building on our completeness findings in Fig.~\ref{fig:ee}, we collect potential hallucinations by identifying objects that frequently appear under different imagination prompts. (3) Interestingly, our detection results align with the original hallucination distribution in Fig.~\ref{fig:intro}, suggesting that context-induced and detected hallucinations mirror those seemingly driven by length, re-validating context is one of the potential factors beneath the iceberg of length.
    \item \textbf{Suppression}: Given the detected potential hallucinations, we can directly suppress their likelihood to mitigate hallucinations. Inspired by contrastive decoding~\cite{damonlpsg2023vcd, kim2024code, wang2024mitigating}, we innovatively treat detected hallucinated objects as contrastive context tokens to their probability in the contrastive branch, thereby reducing their likelihood in the original decoding branches.
\end{itemize}

To sum up, our contributions are as follows:
\begin{itemize}
    \item We are the first to explore the underlying factors beneath the intuitive length-hallucination correlations, and identify context as the potential factor.
    \item We introduce a novel hypothesis based on coherence and completeness, and validate it through statistical analysis, hallucination detection, and suppression.
    \item Our exploration reveals novel insights, including the similarity in image attention patterns of hallucinated objects and the repetition of hallucinations across prompts.
    \item Building on the hypothesis, we propose a novel ``induce-detect-suppress" framework, which re-validates our hypothesis while achieving competitive performance on public benchmarks.
\end{itemize}

\section{Related Work}
\label{sec:rw}

\subsection{Large Vision-Language Models}
The success of large language models (LLMs)~\cite{achiam2023gpt, brown2020language, touvron2023llama, vicuna} establishes the foundation for the development of large visual-language models (LVLMs)~\cite{dai2023instructblip, liu2023llava, zhu2023minigpt, bai2023qwen}. Recent approaches typically adopt a unified framework, where a pre-trained visual encoder extracts visual features, which are then mapped to the LLM embedding space via either linear layers~\cite{liu2023llava, chen2024internvl} or Q-Former~\cite{bai2023qwen, zhu2023minigpt, dai2023instructblip}, and subsequently processed with text inputs. While LVLMs demonstrates remarkable capabilities in visual understanding~\cite{chen2015microsoft, plummer2015flickr30k, schwenk2022okvqa,VQA,hudson2019gqa, yang2019cross-modal, Yang2020Propagating, 9010701, 9710131, 8999516, 9577319, 9408401, Shi_2023_EdaDet, Dai_2024_Curriculum, tang2023temporal, Tang_2023_Contrastive, tang2023cotdet, yang2020graph-structured} and reasoning tasks~\cite{johnson2017clevr,zellers2019recognition,lu2022learn} through supervised fine-tuning~\cite{liu2023llava, huang2025mvtokenflow, he2025vton360highfidelityvirtual, DreamFace, Liang_2024_OMG}, hallucinations remains a prominent challenge~\cite{rohrbach2018object,Li-hallucination-2023,leng2024mitigating,zhou2023analyzing}. Existing studies~\cite{tong2024eyes, fu2025blink, kamath2023s, ICLR2025_3e9a55f6} on the internal mechanisms of LVLMs have yet to provide a thorough explanation of the nature of hallucinations, particularly in long-form responses. This work sheds light on hallucinations in long-form generation in LVLMs.

\subsection{Hallucinations in LVLMs}

Unlike hallucination in LLMs, which refers to the generation of factually incorrect or meaningless content, hallucinations in LVLMs are more concerned with discrepancies between the generated content and the provided visual inputs. Early studies~\cite{rohrbach2018object,Li-hallucination-2023} adapt the definition of hallucinations from the captioning task to the context of LVLMs. Subsequent research~\cite{zhou2023analyzing,damonlpsg2023vcd, liu2023mitigating, huang2024opera} conduct preliminary analyses of hallucinations, investigating factors such as language priors~\cite{damonlpsg2023vcd, liu2023mitigating}, co-occurrence patterns~\cite{zhou2023analyzing, damonlpsg2023vcd}, uncertainty~\cite{zhou2023analyzing}, and positional dependencies~\cite{zhou2023analyzing}.

Several approaches~\cite{zhou2023analyzing,damonlpsg2023vcd, liu2023mitigating, huang2024opera,jiang2024hallucination, zhang2024reflective, kim2024code, wang2024mitigating, liu2024paying, yu2024hallucidoctor} are proposed to mitigate hallucinations in LVLMs through training. These methods include curating high-quality training datasets~\cite{zhou2023analyzing}, integrating specialized contrastive training signals~\cite{jiang2024hallucination}, and employing revisor models designed to correct hallucinated outputs~\cite{liu2023mitigating, yu2024hallucidoctor}. In contrast, other studies~\cite{damonlpsg2023vcd, kim2024code, liu2024paying, huang2024opera, wang2024mitigating} explore training-free strategies as alternatives to resource-intensive training approaches. VCD~\cite{damonlpsg2023vcd} introduces the contrastive decoding (CD)~\cite{li2022contrastive} method to suppress hallucinations, gaining significant attention in the field. Subsequent methods~\cite{wang2024mitigating,liu2024paying,kim2024code} further design various contrastive conditions to induce hallucinations from new perspectives. Additionally, OPERA~\cite{huang2024opera} identifies the overreliance on knowledge aggregation positions within the text attention mechanism as a key cause of hallucinations and suggests a rollback strategy to address this issue.  Furthermore, PAI~\cite{liu2024paying} strengthens the impact of image attention on model outputs, effectively reducing hallucinations.

\section{Is Context a Deeper Underlying Factor?}
\label{sec:explore}

In this section, we conduct exploratory experiments to investigate the underlying factor influencing hallucination beyond generation length. We first introduce PoScore to represent hallucination positions and reproduce the widely recognized phenomenon that hallucinations tend to occur in longer responses (Sec.~\ref{sec:explore_length}). Subsequently, we modify either image or text context and analyze their effects on hallucination distribution, thereby identifying context as a potential underlying factor (Sec.~\ref{sec:explore_context}).

\noindent\textbf{Default Experimental Settings.}
Our default experimental setup (in Sec.~\ref{sec:explore} and Sec.~\ref{sec:analysis}) evaluates the LLaVA v1.5 7B~\cite{liu2023llava}, Qwen VL Chat~\cite{bai2023qwen}, and MiniGPT-4~\cite{zhu2023minigpt} on a randomly sampled set of 500 COCO~\cite{cocodataset} images for statistical analysis. Additional experimental details are presented in Appendix~\ref{sec:appendix_statistical}.

\subsection{Hallucinations Linked to Length.}
\label{sec:explore_length}

When leveraging LVLMs for dialogue or question-answering tasks, a notable phenomenon is that hallucinations tend to occur more frequently in the later positions of the response. To quantitatively analyze this phenomenon, we define the relative position score for each generated object as follows, consistent with previous work~\cite{zhou2023analyzing}:
\begin{equation}
    \text{PoScore}_{s,i} = \frac{\mathrm{Index}(o_{s,i})}{N_s}
\end{equation}

\noindent where $o_{s,i}$ denotes the $i^{th}$ object in the response of the $s^{th}$ sample, and $N_s$ represents the length of the $s^{th}$ sample. We visualize the PoScore distributions for hallucinated and non-hallucinated objects for the LLaVA model in Fig.~\ref{fig:intro}, with additional results from other models provided in Fig.~\ref{fig:intro_app} in Appendix.
The results reveal a marked increase in the frequency of hallucinations as the response lengthens, aligning with findings from previous studies~\cite{zhou2023analyzing, wei2024toward}.

\begin{figure}[t!]
    \centering
    \includegraphics[width=1.\linewidth]{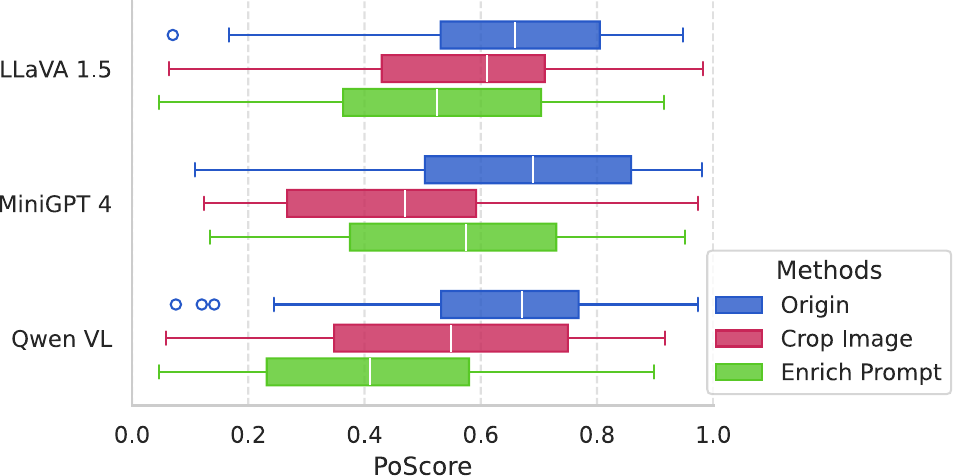}
    \vspace{-6mm}
    \caption{Statistical analysis of hallucination positions under context modifications. Both cropping the image and enriching the prompt lead to earlier hallucination occurrences.}
    \label{fig:manipulate}
    \vspace{-5mm}
\end{figure}

\subsection{Hallucinations Beyond Length.}
\label{sec:explore_context}

Moving beyond these prior observations, we delve deeper by posing a critical question: \textit{Is the increased hallucination merely a result of the cumulative errors due to length itself, or does it arise from a deeper underlying mechanism?} In light of the critical role that context plays in free-form responses, we design the following two context modification strategies and analyze the changes in hallucination positions ($\text{PoScore}$) to investigate the effect of context:

\begin{itemize}
    \item \textbf{Crop the image input} into centered squares, retaining approximately one-third of the original area, and re-annotate accordingly.
    \item \textbf{Enrich the text input} by adding two sentences that describe the image, and then prompt to describe other details.
\end{itemize}

The results in Fig.~\ref{fig:manipulate} show that hallucinations tend to occur earlier in the generation process across both settings, challenging the widely held belief that they are more likely to appear in the later stages. These findings underscore the complexity of hallucinations, revealing that context plays a significant role in their occurrence, rather than attributing them solely to generation length.

\section{Coherence and Completeness}
\label{sec:analysis}

This section delve into the mechanisms through which context influences hallucinations by employing a hypothesis-verification framework. Our analysis focuses on two key aspects: contextual coherence (Sec.~\ref{sec:analysis_repetition}) and contextual completeness (Sec.~\ref{sec:analysis_expansion}). Finally, we link back to text and image manipulation experiments in Sec.~\ref{sec:explore_context}, providing explanations with these factors (Sec.~\ref{sec:analysis_back}).

\begin{figure}[t!]
    \centering
    \includegraphics[width=1.\linewidth]{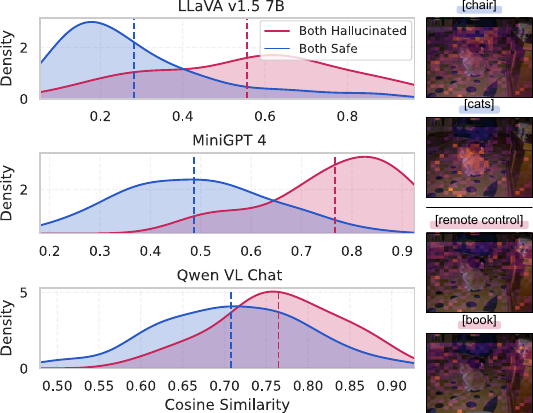}
    \vspace{-4.5mm}
    \caption{Statistical analysis related to contextual coherence. Within the same caption, hallucinated object pairs exhibit higher attention similarity scores than non-hallucinated pairs.}
    \label{fig:compare_attn}
    \vspace{-5mm}
\end{figure}

\subsection{Coherence: Avoidance of Internal Repetition}
\label{sec:analysis_repetition}

Contextual coherence drives the model to maintain consistency with previous outputs while avoiding redundant repetition of both the input and prior content. Based on this, we propose and validate a hypothesis on hallucination occurrence.

\noindent \textbf{Hypothesis.} The two aspects of contextual coherence in image attention are conflicting: attention is required to focus on relevant regions for consistency with previous outputs, while also shifting to new areas to avoid repetition. This tension leads to dispersed attention and hallucinations.

\noindent \textbf{Experimental settings.} To validate our hypothesis, we analyze both individual attention and pairwise attention comparisons. Specifically, we analyze the image attention maps of hallucinated objects $\mathcal{H}$ and non-hallucinated objects $\mathcal{N}$, with representative results shown in Fig.~\ref{fig:compare_attn} (right). Additionally, we quantify the intra-set attention similarity of objects within the same response, denoted by $S_{\mathcal{H}}$ and $S_{\mathcal{N}}$, as follows:
\begin{equation}
    \begin{aligned}
         & S_{\mathcal{H}} = \{ \mathrm{sim}(A_{s,i}, A_{s,j}) \mid o_{s,i},o_{s,j} \in \mathcal{H} \}, \\
         & S_{\mathcal{N}} = \{ \mathrm{sim}(A_{s,i}, A_{s,j}) \mid o_{s,i},o_{s,j} \in \mathcal{N} \}
    \end{aligned}
\end{equation}

\noindent where $A_{s,i}$ and $A_{s,j}$ represent the image attention maps of the $i^{th}$ and $j^{th}$ objects in the response for the $s^{th}$ image, and \(\mathrm{sim}(\cdot, \cdot)\) denotes the cosine similarity function. Fig.~\ref{fig:compare_attn} (left) illustrates the distributions of $S_{\mathcal{H}}$ and $S_{\mathcal{N}}$.

\noindent \textbf{Results.}
Qualitative analysis (right panel of Fig.~\ref{fig:compare_attn}) indicates that when the model successfully identifies real objects, it concentrates on the relevant regions. Conversely, if the model fails to recognize a novel object, its attention disperses and distracting information, leading to hallucinations. Quantitative results (left panel of Fig.~\ref{fig:compare_attn}) show a clear difference between the distributions of $S_{\mathcal{H}}$ and $S_{\mathcal{N}}$. Specifically, hallucinated objects exhibit higher attention similarity, while real objects show lower values. This further indicates that hallucinated objects typically manifest diffuse, noisy attention patterns, making attention similarity a robust metric for their detection.

\begin{figure}[tbp!]
    \centering
    \includegraphics[width=1.\linewidth]{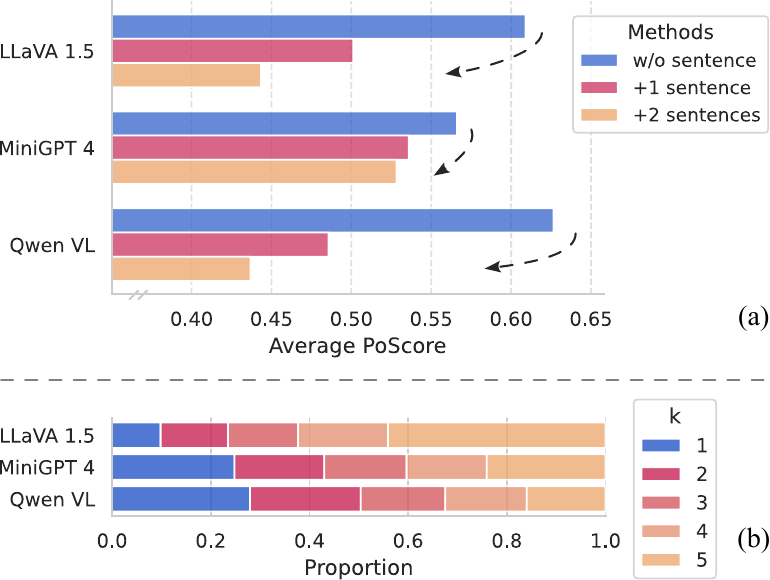}
    \vspace{-6mm}
    \caption{
        Statistical analysis related to contextual completeness: (a) Hallucination positions shift progressively earlier as more image information is included in the prompts. (b) Similar hallucinations consistently recur across varied prompts for the same image.
    }
    \label{fig:ee}
    \vspace{-4mm}
\end{figure}

\subsection{Completeness: External Extrapolation}
\label{sec:analysis_expansion}

Contextual completeness comprises two key dimensions: the informational dimension, which demands a thorough and comprehensive response, and the structural dimension, which ensures the response is logically coherent and grammatically sound. Building on this, we propose the following hypotheses regarding the occurrence mechanism and inherent tendency of hallucination.

\noindent \textbf{Hypothesis.} (a) Occurrence: When a response includes correctly identified objects but remains incomplete in informative or structural aspect, the model compensates by expanding imagined details, i.e., hallucinations. (b) Tendency: These hallucinations from external extrapolation rely on multimodal context, particularly visual inputs.

\noindent \textbf{Experimental settings.} We conduct two separate experiments for validation as follows:

\noindent (a) We validate the role of completeness by analyzing its correlation with hallucination positions. Specifically, we extend text manipulation experiment in Sec.~\ref{sec:explore_context} by incrementally adding image descriptions to the prompt and visualizing the average PoScore in Fig.~\ref{fig:ee}(a).

\noindent (b) We further investigate the consistency and image-related properties of hallucinated objects across different prompts. Specifically, we apply five prompts to each image and compute the proportion of repeated hallucinated objects.
Formally, let $\mathcal{H}_{s_k}$ represent the set of hallucinated objects generated by the $k^{th}$ prompt for the $s^{th}$ sample, with the complete hallucination set given by $\mathcal{H}_s = \bigcup_{k=1}^5 \mathcal{H}_{s_k}$. We count the occurrence of each hallucinated object $h \in \mathcal{H}_{s}$ as $c_s(h) = \sum_{k=1}^{5} \mathds{1}(h \in \mathcal{H}_{s_k})$, where $\mathds{1}$ is the indicator function. Then we calculate $N(k)$, the number of hallucinated objects that appear $k\in [1,2,3,4,5]$ times over all samples, along with its proportion $R(k)$ shown in Fig.~\ref{fig:ee}(b):
\begin{equation}
    \begin{aligned}
        N(k) & = \sum_s\sum_{h\in\mathcal{H}_s}{k\cdot\mathds{1}(c_s(h)=k)}, \\
        R(k) & = \frac{{N(k)}}{\sum_{k=1}^5{N(k)}}
    \end{aligned}
\end{equation}

\noindent \textbf{Results.}
(a) The results in Fig.~\ref{fig:ee}(a) indicate that as more enriched sentences are incorporated, leading to a more comprehensive context, hallucinations occur at earlier positions. This is because the diminishing content available for generation makes it increasingly challenging for LVLMs to accurately identify details for a complete and coherent response. (b) The proportion presented in Fig.~\ref{fig:ee}(b) demonstrate that all models exhibit a high degree of repetitiveness in hallucinated objects, with objects appearing in only one response accounting for merely 30\% on average. Given the variations in both questions and preceding responses, the repeated hallucinated objects are often closely tied to the image context, aligning with our qualitative analysis in Appendix~\ref{sec:appendix_visualization}.

\subsection{Link Back to Phenomenon in Sec.~\ref{sec:explore_context}}
\label{sec:analysis_back}

\noindent \textbf{Explaining Text Manipulation Experiments.}
Revisiting the text manipulation experiments, we find that contextual coherence and completeness provides an intuitive explanation for this behavior. When additional descriptions of real objects are incorporated, the model tend to avoid redundancy and maintain coherence, thereby reducing the number of objects to describe. Consequently, the model turns to uncertain or unverified objects more quickly to ensure completeness, leading to earlier hallucinations.

\noindent \textbf{Explaining Image Manipulation Experiments.}
Contextual completeness offers a compelling explanation for the image manipulation experiments. Similarly, cropping images systematically reduces the number of recognizable objects, forcing the model to hallucinate earlier in order to maintain contextual completeness.

\begin{figure*}[!t]
    \centering
    \includegraphics[width=\linewidth]{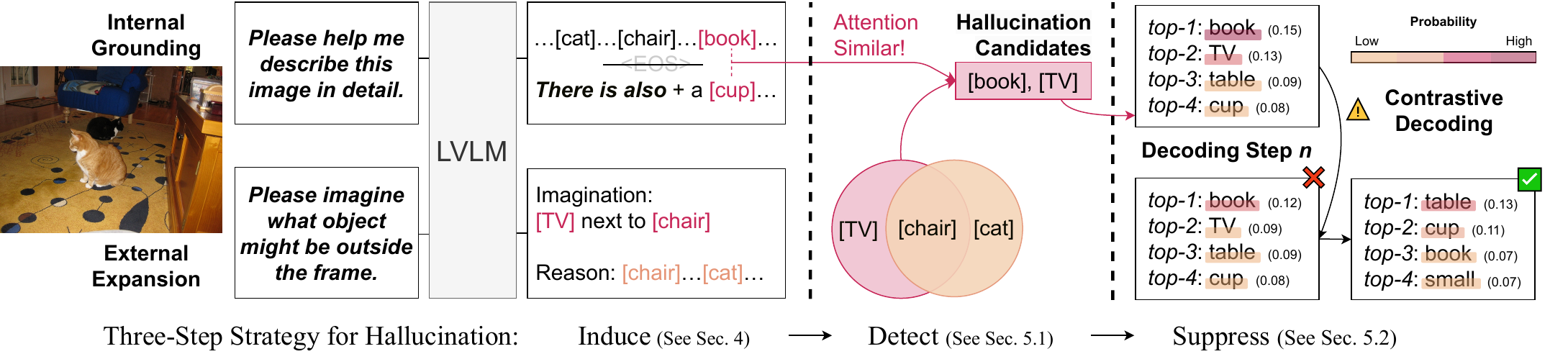}
    \caption{
        \textbf{Overview of HalTrapper:} It consists of two branches leveraging \textit{coherence} and \textit{completeness} insights. One generates captions with an appended ``There is also'' prompt to induce potential hallucinated objects, detected via high attention similarity between caption and induced tokens. The other prompts the LVLM to imagine surrounded content beyond the image to identify consistent hallucinations. With detected hallucinated objects, HalTrapper further suppresses hallucinations through Contrastive Contextual Decoding.
    }
    \label{fig:pipeline}
    \vspace{-4mm}
\end{figure*}

\section{Re-Validation via Detection and Suppression}
\label{sec:method}

To rigorously validate our hypothesis, we extend the findings from Section~\ref{sec:analysis} to practical application of hallucination detection and suppression. Specifically, we propose HalTrapper, which introduces a novel ``induce–detect-suppress" strategy (see Fig.~\ref{fig:pipeline}). The induce–detect stages leverage Internal Grounding (IG) and External Expansion (EE) techniques for hallucination detection (Sec.~\ref{sec:induction}), and can be easily adapted with Contrastive Contextual Decoding (CCD) for suppression (Sec.~\ref{sec:suppression}).

\subsection{Hallucination Induction-Detection}
\label{sec:induction}

\subsubsection{Internal Grounding}
\label{sec:internal}

In Sec.~\ref{sec:analysis_repetition}, we demonstrate that the attention similarity between paired objects serves as an effective indicator for distinguishing hallucinated pairs from non-hallucinated ones. Building on this insight, we propose the Internal Grounding (IG) method, which adopts an \textit{induce-then-detect} paradigm to detect hallucinated objects in model responses.

\noindent \textbf{Induction.}
A key component of IG is the selection of reference objects, which serve as anchors for similarity computation. Instead of using naturally generated objects, we induce the model to generate additional objects following the initial response, which are more prone to hallucination. Specifically, given an input image and the model's initial response, we replace the EOS token in the generated output with an additional cue, ``\textit{There is also}". Since the initial responses inherently covers a considerable number of the identified objects, when completeness is compromised, the model tends to externally extrapolate to compensate, thereby restoring completeness (see Sec.~\ref{sec:analysis_expansion}). The resulting object serves as the reference object and is denoted as $o^{ref}_{s}$ for the $s^{th}$ sample.

\noindent \textbf{Detection.}
We then compute the attention similarity scores $\text{IGScore}$ between the induced hallucinated objects $o^{ref}_{s}$ and the preceding objects, filtering out potential hallucination candidates $S_{IG}$ with high similarity:
\begin{equation}
    \begin{aligned}
        \text{IGScore}_{s,i} & = \mathrm{sim}(A^{ref}_{s}, A_{s,i})                    \\
        S_{IG}               & = \{ o_{s,i} \mid \text{IGScore}_{s,i} > \theta_{IG} \}
    \end{aligned}
\end{equation}
where $\theta_{IG}$ denotes the threshold. Notably, the proposed method remains robust even when the reference object is real, as the similarity scores between non-hallucinated objects are typically low, effectively preventing real objects from being misclassified as hallucinations.

\subsubsection{External Expansion}
\label{sec:external}

Another observation is that hallucinated objects exhibit consistency across identical visual inputs (Sec.~\ref{sec:analysis_expansion}). Based on this property, we propose the External Expansion (EE) method, explicitly \textit{inducing} the imagination related to the image, treating them as \textit{detected} potential hallucinations.

\noindent\textbf{Induction.}
Considering that hallucinations from external extrapolation rely on image context, we first prompt with ``\textit{Please imagine what object might be outside the frame}" to induce image-related associations and capture potential hallucinations. However, directly extracting hallucinated objects from the response leads to false positives, as the model might imagine objects present in the image. To address this, we design a reason-then-imagine prompt to filter out such existing objects (see Appendix~\ref{sec:appendix_ee_prompt}). It explicitly guides the model in distinguishing between recognized objects and imagined ones. Furthermore, it utilizes reliable intermediate steps to enable context-driven reasoning, thereby improving response fidelity.

\noindent\textbf{Detection.}
We introduce EEScore, based on the principle that an object's presence in the imagination set improves the likelihood of it being perceived as a hallucination, while its presence in the reason set reduces this likelihood.
Specifically, we define the imagination set and the reason set at direction $d\in\mathcal{D}$ as $S_{I,d}$ and $S_{R,d}$, respectively. The final set of potential hallucinations is formulated as follows:
\begin{equation}
    \begin{aligned}
        \text{EEScore}_{s,i} & = \sum_{d \in \mathcal{D}}\bigl[\mathds{1}(o_{s,i}\in S_{I,d}) - \mathds{1}(o_{s,i}\in S_{R,d})\bigr] \\
        S_{EE}               & = \{ o_{s,i} \mid \text{EEScore}_{s,i} > \theta_{EE} \}
    \end{aligned}
\end{equation}

Finally, we combine the potential hallucinations detected by the IG and EE methods as follows:
\begin{equation}
    S_{induction} = S_{IG} \cup S_{EE}
\end{equation}

\subsection{Hallucination Suppression}
\label{sec:suppression}

\noindent \textbf{Preliminaries.}
Let $\theta$ denote the parameters of an LVLM. Given an input image $v$ and a text prompt $x$, the model autogressively generates a response $y$ of length $L$. Formally, the decoding process can be formulated as follows:
\begin{equation}
    p_\theta(y|v, x) = \prod_{i=1}^{L} p_\theta(y_i|v,x,y_{<i})
\end{equation}
where $y_i$ and $y_{<i}$ represent the token at position $i$ and preceding tokens before position $i$, respectively, and $p_\theta(y_i|v,x,y_{<i}) \propto \text{exp }\text{logit}_\theta(y_i|v,x,y_{<i})$ denotes the conditional probability distribution of the next token $y_i$ given the preceding tokens $y_{<i}$.

Based on this formulation, we introduce contrastive decoding (CD), originally proposed by~\cite{li2022contrastive}. CD utilizes an amateur model as a contrastive reference to optimize the decoding objectives while maintaining plausibility constraint. Recently, ~\cite{damonlpsg2023vcd, kim2024code, wang2024mitigating} apply CD to LVLMs, leveraging hallucination-amplifying branches as contrastive signals to mitigate hallucinations. Specifically, the CD process, with the new model $\theta'$ as the contrastive branch and all other inputs unchanged, is expressed as follows:
\begin{equation}
    \begin{aligned}
        p_{cd}(y_i|v, x,y_{<i}) = & \Softmax[(1+\alpha)\text{logit}_\theta(y_i|v,x,y_{<i}) \\
                                  & - \alpha \text{logit}_{\theta'}(y_i|v,x,y_{<i})]       \\
    \end{aligned}
\end{equation}
where $p_{\theta'}(x_i|v,x,y_{<i}) \propto \text{exp }\text{logit}_{\theta'}(x_i|v,x,y_{<i})$. It also employs a truncation of the probability distribution following~\cite{damonlpsg2023vcd}.

\noindent\textbf{Contrastive Contextual Decoding (CCD).}
Building on the previously introduced induce-detect stages, a simple CD-based extension CCD enables hallucination suppression. Unlike previous CD methods, CCD explicitly integrates a prior for potential hallucination objects, aiming to reduce their likelihood in response. Specifically, we encode potential hallucinated objects as text tokens, referred to as Contrastive Contextual Tokens (CCT) $ x_{cct}$. We then concatenate CCT with the image input to construct a contrastive branch, with model parameters and other inputs unchanged. The CCD process can be formally expressed as follows:
\begin{equation}
    p_{ccd}(y_i|v, x,y_{<i}) = \prod_{i=1}^{L} p_{ccd}(y_i|v, x_{cct},x,y_{<i})
\end{equation}

We then detail the modifications applied to the CD process as follows:
\begin{equation}
    \begin{aligned}
        p_{ccd} & (y_i|v,x_{cct},x,y_{<i}) =                              \\
                & \Softmax[(1+\alpha)\text{logit}_\theta(y_i|v,x,y_{<i})  \\
                & - \alpha \text{logit}_{\theta}(y_i|v,x,x_{cct},y_{<i})] \\
    \end{aligned}
\end{equation}

\begin{table}[t!]
    \centering
    \small
    \renewcommand\arraystretch{0.5}
    \resizebox{\linewidth}{!}{
        \begin{tabular}{ll|cc|cc}
            \toprule
            \textbf{Model} & \textbf{Metric}   & AUROC         & $\text{TPR}_{5\%\text{FPR}}$ & F1$_\text{max}$ & Acc.          \\
            \cmidrule(lr){1-6}
            \multirow{6}{*}{LLaVA v1.5}
                           & PoScore           & $70.7$        & $4.3$                        & $38.3$          & $70.7$        \\
                           & Top Logit         & $64.0$        & $13.0$                       & $32.2$          & $61.9$        \\
                           & Logits' Entropy   & $67.7$        & $16.6$                       & $36.6$          & $71.4$        \\
                           & Image Attn. Ratio & $44.9$        & $6.0$                        & $27.3$          & $32.0$        \\
            \cmidrule(lr){2-6}
                           & IG Score          & \textbf{82.3} & \textbf{43.3}                & \textbf{54.8}   & \textbf{86.3} \\
                           & EE Score          & $77.5$        & -                            & $46.1$          & $72.9$        \\
            \cmidrule(lr){1-6}
            \multirow{6}{*}{MiniGPT 4}
                           & PoScore           & $70.5$        & $12.2$                       & $35.4$          & $66.2$        \\
                           & Top Logit         & $65.6$        & $22.9$                       & $37.0$          & $76.5$        \\
                           & Logits' Entropy   & $65.5$        & $22.1$                       & $35.3$          & $75.9$        \\
                           & Image Attn. Ratio & $64.3$        & $7.7$                        & $31.9$          & $57.9$        \\
            \cmidrule(lr){2-6}
                           & IG Score          & \textbf{76.6} & \textbf{34.0}                & \textbf{48.6}   & \textbf{80.7} \\
                           & EE Score          & $60.5$        & -                            & $30.0$          & $46.5$        \\
            \cmidrule(lr){1-6}
            \multirow{6}{*}{Qwen VL}
                           & PoScore           & $71.1$        & $4.8$                        & $34.4$          & $65.8$        \\
                           & Top Logit         & $71.5$        & $19.6$                       & $36.1$          & $77.7$        \\
                           & Logits' Entropy   & $70.7$        & $23.3$                       & $36.6$          & $73.9$        \\
                           & Image Attn. Ratio & $57.3$        & $6.8$                        & $26.9$          & $41.4$        \\
            \cmidrule(lr){2-6}
                           & IG Score          & $76.2$        & \textbf{33.3}                & $43.8$          & \textbf{84.6} \\
                           & EE Score          & \textbf{81.3} & -                            & \textbf{46.3}   & $73.0$        \\
            \bottomrule
        \end{tabular}
    }
    \vspace{-2mm}
    \caption{
        Quantitative results for hallucination detection. The best performances within each setting are \textbf{bolded}.
    }
    \label{tab:detection}
    \vspace{-3.4mm}
\end{table}

By treating CCT tokens as complementary to image content, the model naturally increases the likelihood of potential hallucinated objects and their associated terms in the contrastive branch, thereby effectively reducing their occurrence in the final generation.

\section{Experiments}
\label{sec:experiment}

\noindent \textbf{Datasets and Benchmarks.}
To demonstrate the effectiveness of our HalTrapper, we use images from \textbf{COCO}~\cite{cocodataset} and \textbf{AMBER}~\cite{wang2023amber} datasets. Detailed descriptions can be found in the Appendix~\ref{sec:appendix_dataset}.
\begin{figure}[t!]
    \centering
    \includegraphics[width=1.\linewidth]{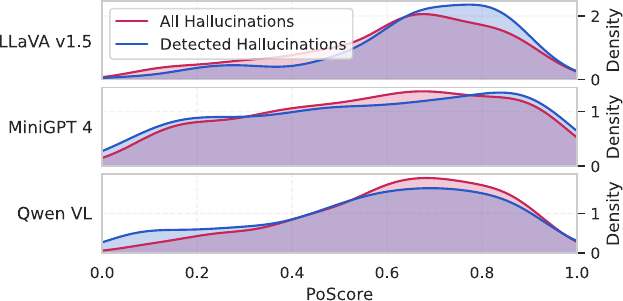}
    \vspace{-5mm}
    \caption{
        Comparison between the positional distribution of hallucinations detected by our method and the overall hallucination distribution, demonstrating a high degree of alignment.
    }
    \label{fig:detect_distribution}
    \vspace{-4mm}
\end{figure}

\noindent \textbf{Base Models.}
We select LLaVA v1.5 7B~\cite{liu2023llava}, MiniGPT-4~\cite{zhu2023minigpt}, and Qwen VL Chat~\cite{bai2023qwen} as our main baselines for our study. We also evaluate more recent models Qwen2 VL 7B~\cite{Qwen2-VL} and Janus Pro 7B~\cite{chen2025janus} on AMBER, which has higher annotation qualities.

\noindent \textbf{Implementation Details.}
For all experiments, the maximum number of newly generated tokens is set to 512. Following prior mainstream studies on CD~\cite{damonlpsg2023vcd,wang2024mitigating}, we adapt $\alpha = 1.0$ and $\beta = 0.1$. See Appendix~\ref{sec:appendix_cct} and~\ref{sec:appendix_hyperparameter} for details on CCT construction and more hyperparameters.

\subsection{Detection}

\noindent\textbf{Metrics.}
Inspired by~\cite{sriramanan2024llmcheck}, we adapt AUROC (Area Under the ROC Curve) and TPR@5\%FPR (the True Positive Rate at 5\% False Positive Rate) as our primary metrics for hallucination detection. AUROC quantifies the model's overall discriminative ability across all classification thresholds, while TPR@5\%FPR is suitable for scenarios with strict requirements on the false positive rate. We also report the F1 Score and Accuracy at the threshold that \textit{maximizes} the F1.

\noindent \textbf{Baseline Methods.}
For each generated object \( o_{s,i} \), we first employ PoScore~\cite{zhou2023analyzing} as a basic metric. We also propose two uncertainty-based metrics: Top Logit and Logits' Entropy. The Top Logit is the maximum value of the logits when generating \( o_{s,i} \), while Logits' Entropy refers to the entropy of the logits at that moment. Additionally, we employ an Attention-based metric called the Image Attention Ratio, defined as the ratio of the model's attention score on the image to its total attention score when generating \( o_{s,i} \).

\begin{table}[t!]
    \centering
    \small
    \setlength\tabcolsep{1.2mm}
    \renewcommand\arraystretch{0.5}

    \resizebox{1.\linewidth}{!}{
        \begin{tabular}{llcccccc}
            \toprule
            \multirow{2}{*}{\textbf{Decoding}} & \multirow{2}{*}{\textbf{Method}} & \multicolumn{6}{c}{\textbf{LLaVA v1.5 7B~\cite{liu2023llava}}}                                                                                                                                                                          \\
            \cmidrule(l{0pt}r{0pt}){3-8}
                                               &                                  & $\mathrm{C_S}$$\downarrow$                                     & $\mathrm{C_I}$$\downarrow$            & Prec.                               & Recall                                   & F1                                  & Len     \\
                    \midrule
            \multirow{5}{*}{Greedy}            & ICD~\cite{wang2024mitigating}    & $51.4$                                                         & $14.7$                                & $73.4$                              & $81.0$                                   & $77.0$                              & $102.1$ \\
                                               & CODE~\cite{kim2024code}          & $50.0$                                                         & $13.7$                                & $75.8$                              & $76.9$                                   & $76.4$                              & $88.3$  \\
                    \cmidrule(l{0pt}r{0pt}){2-8}
                                               & Vanilla                          & $52.2$                                                         & $14.6$                                & $73.7$                              & $80.3$                                   & $76.9$                              & $100.8$ \\
                                               & Ours                             & \textbf{41.6}                                                  & \textbf{11.9}                         & \textbf{78.7}                       & $80.1$                                   & \textbf{79.4}                       & $100.0$ \\
                                               &                                  & \textcolor{iccvblue}{$10.6\downarrow$}                         & \textcolor{iccvblue}{$2.7\downarrow$} & \textcolor{iccvblue}{$5.0\uparrow$} & \textcolor{red!30!gray}{$0.2\downarrow$} & \textcolor{iccvblue}{$2.5\uparrow$} &         \\
                    \midrule
            \multirow{6}{*}{Nucleus}           & VCD~\cite{damonlpsg2023vcd}      & $58.2$                                                         & $16.9$                                & $70.8$                              & $78.8$                                   & $74.6$                              & $103.2$ \\
                                               & ICD~\cite{wang2024mitigating}    & $55.0$                                                         & $16.5$                                & $70.9$                              & $77.9$                                   & $74.2$                              & $102.1$ \\
                                               & CODE                             & $54.2$                                                         & $16.4$                                & $72.3$                              & $76.2$                                   & $74.2$                              & $91.6$  \\
                    \cmidrule(l{0pt}r{0pt}){2-8}
                                               & Vanilla                          & $58.6$                                                         & $18.8$                                & $68.1$                              & $76.4$                                   & $72.0$                              & $105.2$ \\
                                               & Ours                             & \textbf{48.6}                                                  & \textbf{14.5}                         & \textbf{74.6}                       & \textbf{77.7}                            & \textbf{76.1}                       & $100.9$ \\
                                               &                                  & \textcolor{iccvblue}{$10.0\downarrow$}                         & \textcolor{iccvblue}{$4.3\downarrow$} & \textcolor{iccvblue}{$6.5\uparrow$} & \textcolor{iccvblue}{$1.3\downarrow$}    & \textcolor{iccvblue}{$4.1\uparrow$} &         \\
                    \midrule
            \multirow{4}{*}{Beam Search}       & OPERA~\cite{huang2024opera}      & $53.6$                                                         & $15.7$                                & $72.4$                              & $77.6$                                   & $74.9$                              & $98.8$  \\
                    \cmidrule(l{0pt}r{0pt}){2-8}
                                               & Vanilla                          & $55.6$                                                         & $15.8$                                & $72.8$                              & $81.0$                                   & $76.7$                              & $104.2$ \\
                                               & Ours                             & \textbf{45.2}                                                  & \textbf{12.1}                         & \textbf{78.9}                       & \textbf{81.2}                            & \textbf{80.0}                       & $101.8$ \\
                                               &                                  & \textcolor{iccvblue}{$10.4\downarrow$}                         & \textcolor{iccvblue}{$3.7\downarrow$} & \textcolor{iccvblue}{$6.1\uparrow$} & \textcolor{iccvblue}{$0.2\downarrow$}    & \textcolor{iccvblue}{$3.3\uparrow$} &         \\
            \bottomrule
        \end{tabular}
    }
    \vspace{-1.5mm}
    \caption{
        Results on CHAIR. Lower CHAIR$_S$, CHAIR$_I$, and higher precision, recall and F1 indicate fewer hallucinations. The best performances within each setting are \textbf{bolded}.
    }
    \label{tab:chair_llava}
    \vspace{-4.4mm}
\end{table}

\noindent \textbf{Results.}
The quantitative results of hallucination detection are presented in Table~\ref{tab:detection}. As shown, our approach demonstrates significant improvements across all evaluation settings. For IG, in terms of the AUROC metric, our method outperforms the best baseline PoScore by 5\%–12\%. This indicates that our method enhances performance across the entire classification curve. Considering that our IG method originates from within the model, this indicates that the model indeed exhibits significant similar attention pattern in certain hallucination scenarios. Additionally, for TPR@5\%FPR, our method improves by at least 10\% compared to the baselines. This highlights the substantial potential of the EE metric in inducing hallucinations. Given that MiniGPT 4 is trained only on the image interface, its ability to follow instructions is relatively limited, which may account for the lack of improvement in the EE metric.

We also visualized the qualitative results of hallucination positions distribution detected by our method, with the overall distribution of hallucination positions, as shown in Fig.~\ref{fig:detect_distribution}. It demonstrates that our method accurately captures the hallucination distribution, closely aligning with the overall pattern observed in captions. This further indicates that although we claim that our method is designed for long-text scenarios, its effectiveness is not merely dependent on the length of the generated text.  Instead, our approach effectively captures an intrinsic mechanism underlying LVLM hallucinations, which is beyond text length.  Therefore, our study not only validates the applicability of our method but also provides a new perspective for understanding the formation mechanism of LVLM hallucinations.

\subsection{Suppression}

\noindent \textbf{Metrics.}
\textbf{CHAIR}~\cite{rohrbach2018object} is commonly used to quantify hallucinations in model-generated captions based on COCO. Besides CHAIR, we also report several classic metrics, including Precision, Recall, F1, and the average length of the captions. For \textbf{AMBER}~\cite{wang2023amber}, following the approach outlined in their paper, we report CHAIR, Cover, Hal, and Cog. As we primarily focus on long context scenarios, we conduct full evaluations only on its generative subset and reported the results accordingly. We also conduct experiments on POPE and GPT-4o, please refer to the Appendix~\ref{sec:appendix_pope} and~\ref{sec:appendix_gpt}.

\begin{table}[t!]
    \centering
    \small
    \setlength\tabcolsep{1.5mm}
    \renewcommand\arraystretch{0.5}

    \resizebox{1.\linewidth}{!}{
        \begin{tabular}{llcccccc}
            \toprule
            \multirow{2}{*}{\textbf{Decoding}} & \multirow{2}{*}{\textbf{Method}} & \multicolumn{3}{c}{\textbf{MiniGPT 4}~\cite{zhu2023minigpt}} & \multicolumn{3}{c}{\textbf{Qwen VL Chat}~\cite{bai2023qwen}}                                                                                               \\
            \cmidrule(l{0pt}r{1pt}){3-5} \cmidrule(l{1pt}r{0pt}){6-8}
                                               &                                  & $\mathrm{C_S}$$\downarrow$                                   & $\mathrm{C_I}$$\downarrow$                                   & Prec.           & $\mathrm{C_S}$$\downarrow$ & $\mathrm{C_I}$$\downarrow$ & Prec.           \\
                    \midrule
            \multirow{4}{*}{Greedy}            & Vanilla                          & $39.6$                                                       & $14.7$                                                       & $76.6$          & $43.4$                     & $13.5$                     & $75.8$          \\
                                               & ICD~\cite{wang2024mitigating}    & $42.6$                                                       & $14.7$                                                       & $76.3$          & $50.4$                     & $14.4$                     & $73.7$          \\
                                               & CODE~\cite{kim2024code}          & $32.8$                                                       & $13.6$                                                       & $81.2$          & $40.4$                     & $12.5$                     & $78.9$          \\
                                               & Ours                             & $\textbf{28.6}$                                              & $\textbf{10.7}$                                              & $\textbf{83.1}$ & $\textbf{38.6}$            & $\textbf{10.2}$            & $\textbf{80.9}$ \\
                    \midrule
            \multirow{5}{*}{Nucleus}           & Vanilla                          & $37.2$                                                       & $14.6$                                                       & $77.1$          & $44.8$                     & $13.6$                     & $76.3$          \\
                                               & VCD~\cite{damonlpsg2023vcd}      & $39.6$                                                       & $14.9$                                                       & $76.6$          & $47.4$                     & $14.1$                     & $74.3$          \\
                                               & ICD                              & $41.4$                                                       & $14.9$                                                       & $76.1$          & $52.6$                     & $15.0$                     & $73.0$          \\
                                               & CODE~\cite{kim2024code}          & $36.6$                                                       & $14.0$                                                       & $79.5$          & $43.6$                     & $14.5$                     & $75.4$          \\
                                               & Ours                             & $\textbf{29.0}$                                              & $\textbf{11.5}$                                              & $\textbf{82.1}$ & $\textbf{42.4}$            & $\textbf{11.3}$            & $\textbf{79.3}$ \\
                    \midrule
            \multirow{3}{*}{Beam Search}       & Vanilla                          & $38.8$                                                       & $13.8$                                                       & $78.0$          & $41.4$                     & $11.6$                     & $79.0$          \\
                                               & OPERA~\cite{huang2024opera}      & $43.0$                                                       & $14.9$                                                       & $75.8$          & $42.8$                     & $12.5$                     & $76.9$          \\
                                               & Ours                             & $\textbf{37.6}$                                              & $\textbf{13.7}$                                              & $\textbf{78.3}$ & $\textbf{34.2}$            & $\textbf{9.7}$             & $\textbf{82.7}$ \\
            \bottomrule
        \end{tabular}
    }
    \vspace{-2mm}
    \caption{More results on CHAIR with MiniGPT-4 and Qwen VL.
    }
    \label{tab:chair_2}
    \vspace{-1.2mm}
\end{table}

\begin{table}[t]
    \renewcommand\arraystretch{0.5}
    \centering
    \resizebox{1.\linewidth}{!}{%
        \begin{tabular}{lcccc}
            \toprule
            \textbf{Model / Method}           & CHAIR$\downarrow$            & Cover$\uparrow$             & Hal$\downarrow$                & Cog$\downarrow$              \\
            \midrule
            LLaVA v1.5 7B~\cite{liu2023llava} & $11.2$                       & $50.2$                      & $47.9$                         & $4.6$                        \\
            + VCD~\cite{damonlpsg2023vcd}     & $8.9$                        & $51.2$                      & $38.1$                         & $4.4$                        \\
            + ICD~\cite{wang2024mitigating}   & $8.6$                        & $51.1$                      & $37.3$                         & $3.9$                        \\
            + CODE~\cite{kim2024code}         & $9.0$                        & $51.1$                      & $39.5$                         & $4.3$                        \\
            + Ours                            & \res{$8.0$}{3.2$\downarrow$} & \res{$51.5$}{1.3$\uparrow$} & \res{$36.3$}{11.6$\downarrow$} & \res{$3.8$}{0.8$\downarrow$} \\
            \midrule
            Qwen2 VL~\cite{Qwen2-VL}          & $6.6$                        & $71.8$                      & $50.3$                         & $4.6$                        \\
            + VCD                             & $7.3$                        & $70.6$                      & $53.2$                         & $4.6$                        \\
            + ICD                             & $8.2$                        & $74.9$                      & $74.9$                         & $9.1$                        \\
            + CODE                            & $7.6$                        & $71.6$                      & $56.3$                         & $5.1$                        \\
            + Ours                            & \res{$5.6$}{1.0$\downarrow$} & $70.9$                      & \res{$46.1$}{4.2$\downarrow$}  & \res{$3.8$}{0.8$\downarrow$} \\
            \midrule
            Janus Pro 7B~\cite{chen2025janus} & $6.3$                        & $65.6$                      & $37.5$                         & $2.0$                        \\
            + VCD                             & $5.5$                        & $66.2$                      & $32.5$                         & $2.1$                        \\
            + ICD                             & $6.1$                        & $67.1$                      & $36.3$                         & $2.5$                        \\
            + CODE                            & $6.0$                        & $65.3$                      & $33.6$                         & $1.6$                        \\
            + Ours                            & \res{$5.4$}{0.9$\downarrow$} & \res{$66.5$}{0.9$\uparrow$} & \res{$32.7$}{4.8$\downarrow$}  & \res{$1.8$}{0.2$\downarrow$} \\
            \bottomrule
        \end{tabular}
    }
    \vspace{-2mm}
    \caption{Results on AMBER~\cite{wang2023llm} generative task. $\downarrow$ indicates lower is better.}
    \label{tab:amber}
    \vspace{-5mm}
\end{table}

\noindent \textbf{Baseline Methods.}
We compare our HalTrapper with VCD~\cite{damonlpsg2023vcd}, ICD~\cite{wang2024mitigating}, CODE~\cite{kim2024code}, and OPERA~\cite{huang2024opera}.

\noindent \textbf{CHAIR Evaluation.}
As shown in Tables~\ref{tab:chair_llava} and~\ref{tab:chair_2}. HalTrapper significantly reduces CHAIR while maintaining Recall with minimal negative impact. Across all experiments on CHAIR$_S$ and CHAIR$_I$, HalTrapper achieves significant improvements. Notably, in Table~\ref{tab:chair_llava}, our approach consistently improves CHAIR$_S$ by over 10\% and CHAIR$_I$ by 2.5\%. This demonstrates that the hallucination candidates identified by our IG and EE metrics are of high quality, enabling the inclusion of a large number of hallucinated objects while minimizing the presence of non-hallucinated ones. This, in turn, provides validation of the effectiveness of our IG and EE metrics in detecting hallucinations, further highlighting the universality and practical significance of our findings.

\noindent \textbf{AMBER Evaluation.}
As shown in the Table~\ref{tab:amber}, HalTrapper continues to demonstrate performance improvements on latest models.
\noindent \textbf{Ablation Study.}
See Appendix~\ref{sec:appendix_abla} for more details on the ablation study.

\section{Conclusion}
\label{sec:conclusion}

In this paper, we propose a novel method for eliminating hallucinations in Large Vision-Language Models through two mechanisms: external spatial expansion and internal visual grounding. Our HalTrapper introduces a simple, zero-shot hallucination detection and suppression technique that achieves significant improvements across all benchmarks, with no additional training required. Our approach consistently delivers substantial improvements across all benchmarks, validating its effectiveness.

\noindent\textbf{Acknowledgment.} This work is supported by the National Natural Science Foundation of China (No.62206174).

{
    \small
    \bibliographystyle{ieeenat_fullname}
    \bibliography{main}
}

\appendix
\maketitlesupplementary
\setcounter{page}{1}

\noindent This supplementary material provides further details on our findings, the specific prompts and configurations used in our experiments, additional quantitative and qualitative results, and a discussion of limitations. Specifically, we first provide supplementary experimental settings used in our analysis experiments (Sec.~\ref{sec:appendix_statistical}). Next, we present complementary results to support our analysis (Sec.~\ref{sec:appendix_analysis}). We then describe further implementation details and experimental setups for the experiments in the main paper (Sec.~\ref{sec:appendix_impli}). Additionally, we conduct ablation studies and evaluate HalTrapper on additional benchmarks to further validate its effectiveness (Sec.~\ref{sec:appendix_exp}). We also include visualizations to aid comparison and provide a clearer understanding of HalTrapper (Sec.~\ref{sec:appendix_visualization}). Finally, we provide a discussion of the limitations of our work (Sec.~\ref{sec:appendix_limitations}).

\section{Supplementary Details on Exploratory Experiments and Analyses}
\label{sec:appendix_statistical}

\subsection{Settings for Hallucinations Beyond Length}

For the experiment of modifying image and text context (Sec.~\ref{sec:explore_context}), since the image cropping experiment requires manual re-annotation of cropped part, we randomly sample 50 images from COCO dataset for this experiment.

\subsection{Prompt Design for Completeness}

In Fig.~\ref{fig:ee}(a) of the paper, we demonstrate that the model is more prone to hallucinations when its content is incomplete by adjusting the amount of textual context inserted into the model. To eliminate the influence of length, we designed prompts of different lengths for different groups, ensuring that the total number of sentences in each prompt remains consistent (4 here). Although the prompt lengths varied in our design, we endeavored to maintain consistency in the information contained within them as much as possible. Below are the specific prompts we used, where \{\} are placeholders for sentences to be inserted:

\begin{itemize}
    \item Group \textbf{w/o sentence}: \textit{Please help me describe this image in detail. I'd like to hear more about it, even if it's just small things. Anything you can say about it would be useful in some way. It doesn't have to be important, just whatever comes to mind.}
    \item Group \textbf{+1 sentence}: \textit{I already know that \{\} Could you describe any other details of the image for me? It doesn't have to be anything specific, just whatever else you can say about it. Even if it seems unimportant, it might still be worth mentioning.}
    \item Group \textbf{+2 sentences}: \textit{I already know that \{\} Could you describe any other details of the image for me? Maybe there's something that hasn't been mentioned yet, or just anything that comes to mind.}
\end{itemize}

\begin{figure}[t!]
    \centering
    \includegraphics[width=1.\linewidth]{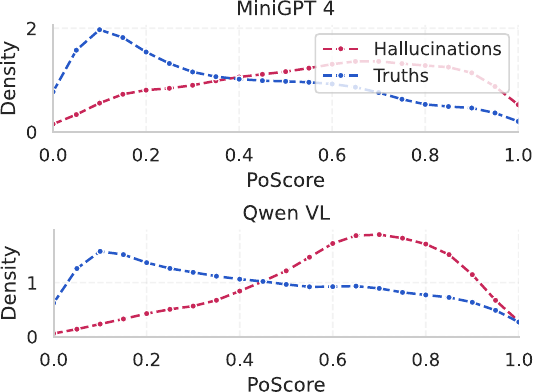}
    \caption{Distribution of hallucinated and non-hallucinated object positions in captions generated by different models.}
    \label{fig:intro_app}
    \vspace{-4mm}
\end{figure}

\section{Additional Results for Exploratory Experiments and Analysis}
\label{sec:appendix_analysis}

\subsection{Additional Baseline Results for Hallucinations Linked to Length}
We conduct the same experiments as Sec.~\ref{sec:explore_length} on Qwen VL Chat and MiniGPT-4. Results are shown in Fig.~\ref{fig:intro_app}. The results demonstrate that Qwen VL Chat and MiniGPT-4 also show a pronounced tendency for increased hallucinations with longer input contexts.

\subsection{Qualitative Support for Statistical Analysis}

In the main paper, we conduct a series of statistical experiments to demonstrate that hallucinations in LVLMs are not solely related to input length, but also influenced by coherence and completeness. To facilitate understanding, we provide qualitative examples of the experiments here.

\begin{figure}[t]
    \centering
    \includegraphics[width=1.\linewidth]{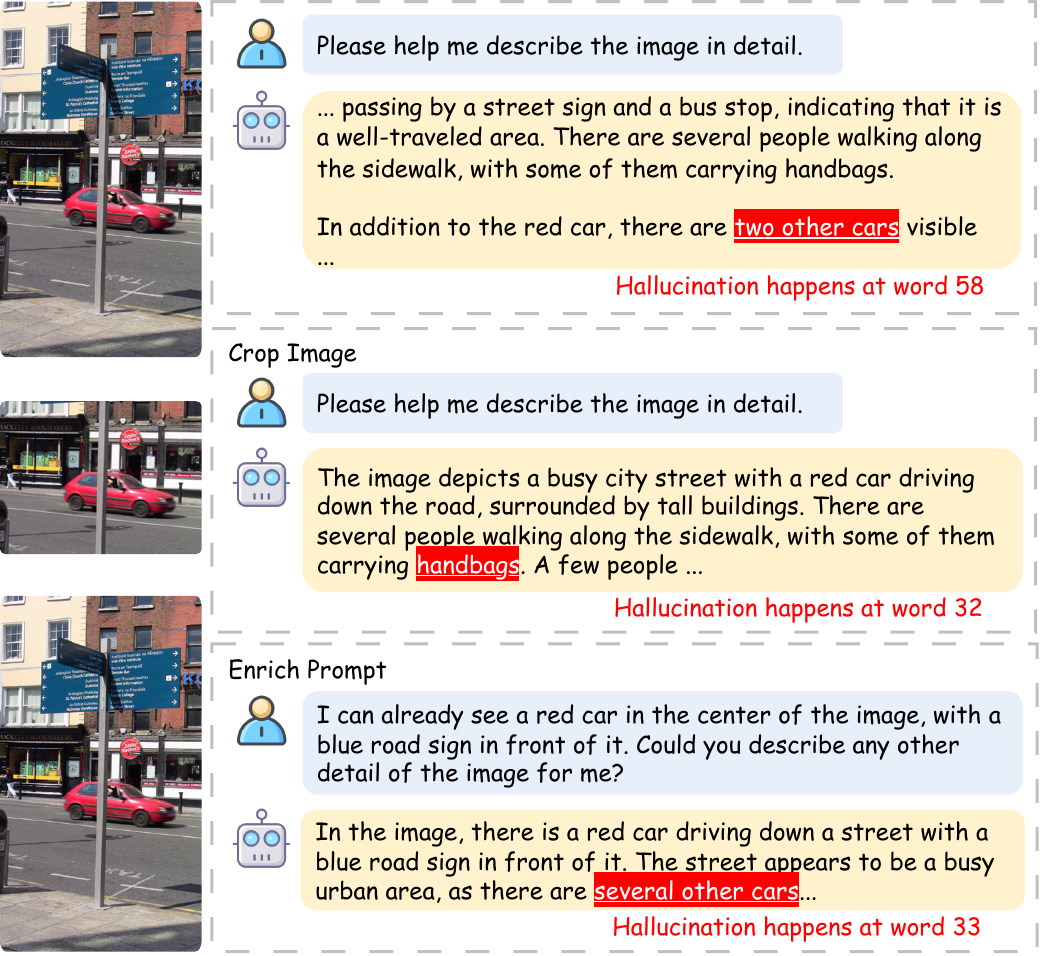}
    \caption{Illustrative example of hallucination positions under context modifications, corresponding to the mechanism shown in Fig.~\ref{fig:manipulate}. Both cropping the image and enriching the prompt lead to earlier hallucination occurrences. Hallucinations are highlighted in \textcolor{red}{\textbf{red}.}}
    \label{fig:qf2}
\end{figure}

\noindent \textbf{Illustrations for Hallucination Beyond to Length.} Fig.~\ref{fig:qf2} presents an example from the experiment described in Fig.~\ref{fig:manipulate} of the main text. It can be observed that both cropping the image and enriching the prompt lead to earlier occurrences of hallucinations.

\begin{figure}[t]
    \centering
    \includegraphics[width=1.\linewidth]{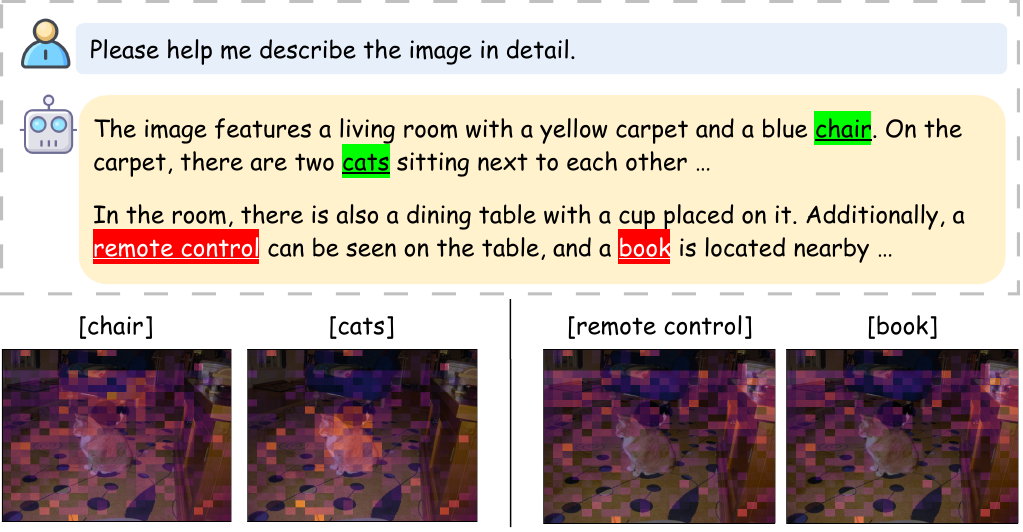}
    \caption{Illustrative example of attention similarity between hallucinated and non-hallucinated object pairs within the same caption, corresponding to the mechanism shown in Fig.~\ref{fig:compare_attn}. Hallucinated pairs exhibit significantly higher attention similarity scores. Hallucinations are highlighted in \textcolor{red}{\textbf{red}.}}
    \label{fig:qf3}
\end{figure}

\noindent \textbf{Illustrations for Coherence Analysis.} Fig.~\ref{fig:qf3} supplements the visualization on the right side of Fig.~\ref{fig:compare_attn} with a complete example, illustrating that hallucinated pairs exhibit significantly higher attention similarity scores.

\begin{figure}[t]
    \centering
    \includegraphics[width=1.\linewidth]{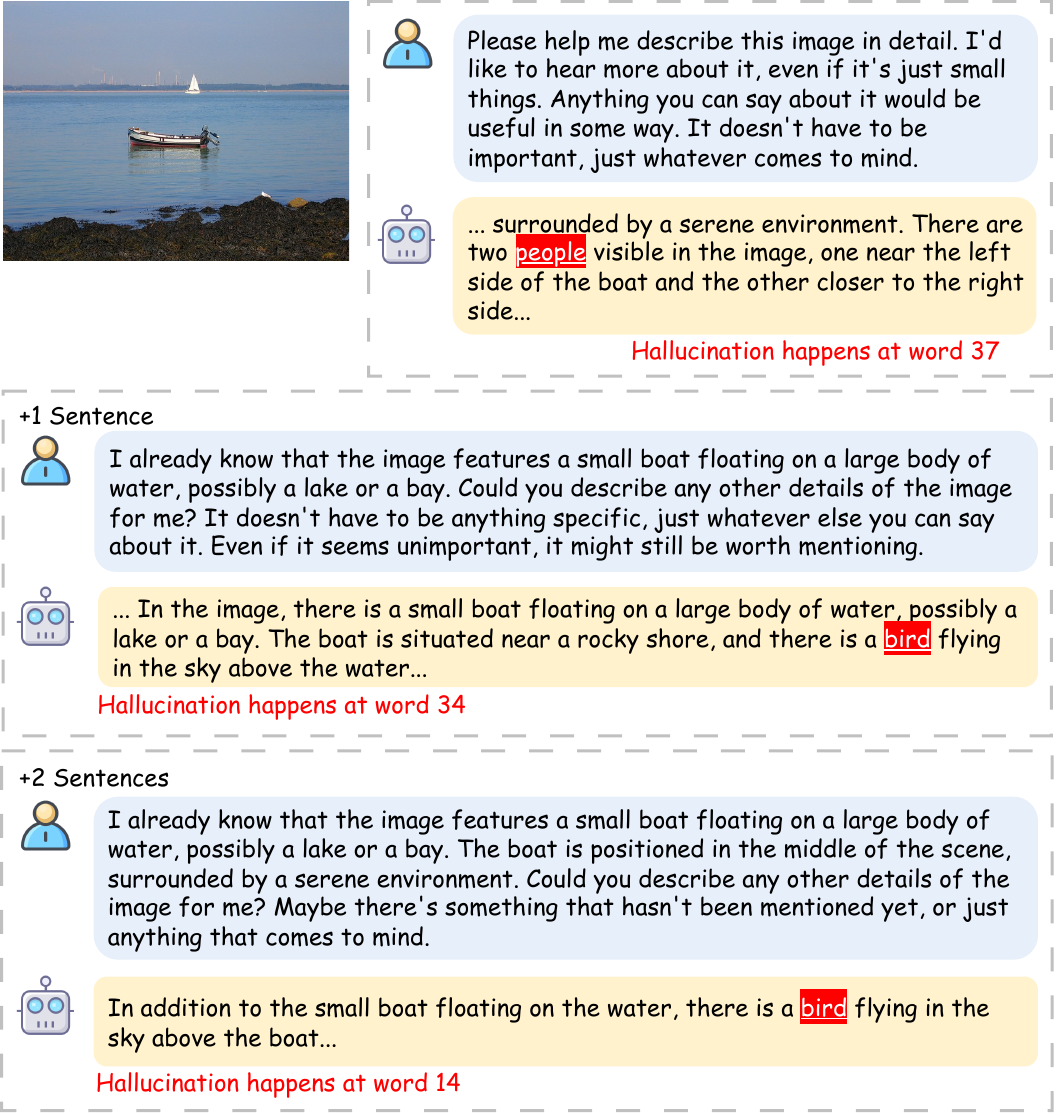}
    \caption{Illustrative example of the positions where hallucinations occur as the visual information in prompts gradually increases, while the prompt length remains similar, corresponding to the mechanism shown in Fig.~\ref{fig:ee}(a). Hallucinations tend to appear earlier when more visual context is included. Hallucinations are highlighted in \textcolor{red}{\textbf{red}}.}
    \label{fig:qf4a}
\end{figure}

\begin{figure}[t]
    \centering
    \includegraphics[width=1.\linewidth]{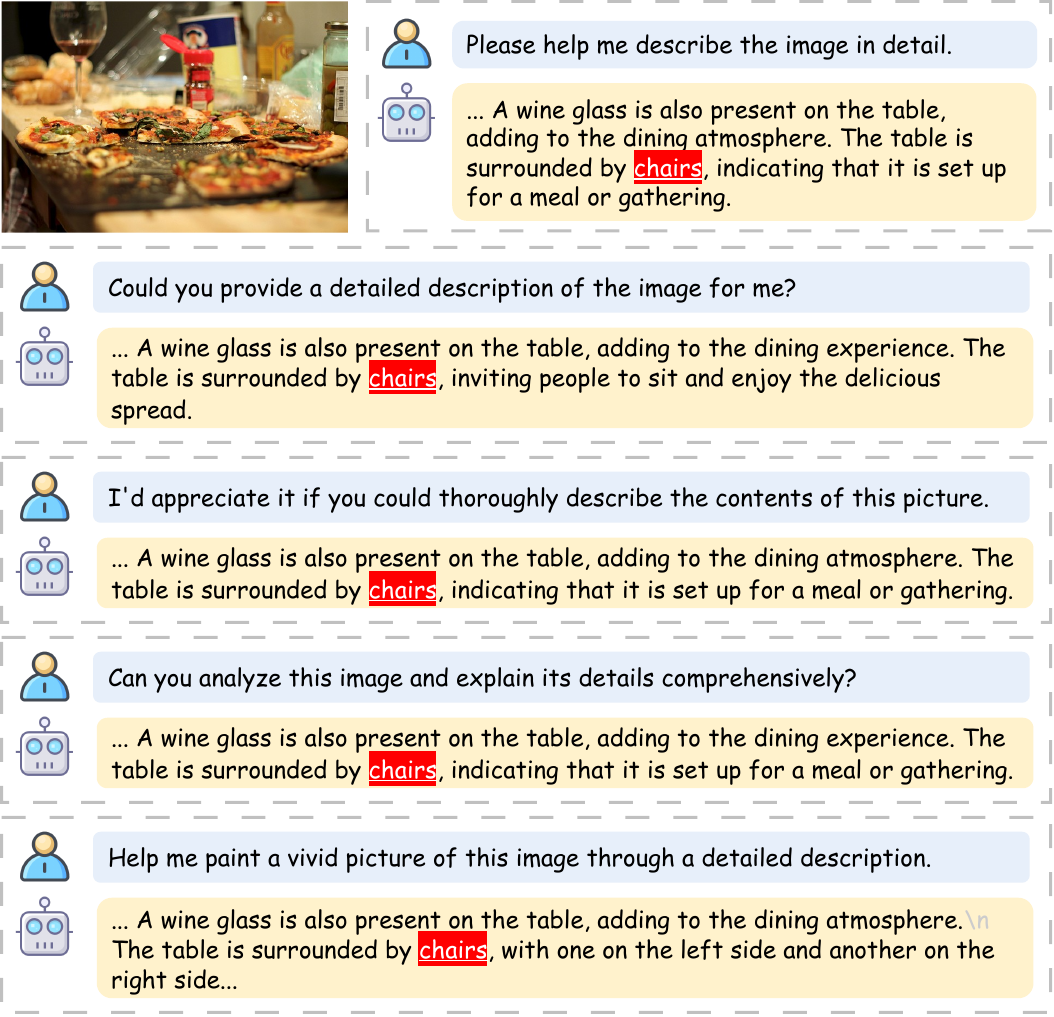}
    \caption{Illustrative example of the repetition frequency distribution of hallucinated objects across different prompts for the same image, corresponding to the mechanism shown in Fig.~\ref{fig:ee}(b). Similar hallucinations consistently appear despite changes in prompts. Hallucinations are highlighted in \textcolor{red}{\textbf{red}.}}
    \label{fig:qf4b}
\end{figure}

\noindent \textbf{Illustrations for Completeness Analyses.} Fig.~\ref{fig:qf4a} and Fig.~\ref{fig:qf4b} visualize specific examples from the two experiments shown in Fig.~\ref{fig:ee}(a) and (b) of the main text, respectively. Fig.~\ref{fig:qf4a} further demonstrates that hallucinations tend to appear earlier when more visual context is included, while Fig.~\ref{fig:qf4b} shows that similar hallucinations consistently emerge despite variations in prompts.

\section{Detailed Implementation and Experimental Setup}
\label{sec:appendix_impli}

\begin{table}[t!]
    \centering
    \small
    \setlength\tabcolsep{4mm}
    \renewcommand\arraystretch{1}
    \resizebox{.9\linewidth}{!}{
        \begin{tabular}{lcccc}
            \toprule
            \textbf{Model} & $\theta_{IG}$ & $\theta_{EE}$ & $N$ & $T_{sep}$     \\
            \midrule
            LLaVA v1.5 7B  & 0.75          & 1             & 10  & `\texttt{ }'  \\
            MiniGPT 4      & 0.75          & 0             & 10  & `\texttt{, }' \\
            Qwen VL Chat   & 0.85          & 0             & 5   & `\texttt{ }'  \\
            Qwen2 VL 7B    & 0.75          & 1             & 5   & `\texttt{ }'  \\
            Janus Pro 7B   & 0.75          & 1             & 5   & `\texttt{ }'  \\
            \bottomrule
        \end{tabular}
    }
    \caption{Parameters used for hallucination suppression.}
    \label{tab:parameters}
    \vspace{-4mm}
\end{table}

\subsection{Details of Datasets and Benchmarks.}
\label{sec:appendix_dataset}

\noindent \textbf{COCO}~\cite{cocodataset}, the Common Objects in Context dataset is widely used in computer vision, providing detailed annotations for 80 object categories and serving as a valuable resource for evaluating hallucination detection and suppression.

\noindent \textbf{AMBER}~\cite{wang2023amber}, an LLM-free multi-dimensional benchmark, is also specifically designed to assess hallucinations in LVLMs. With 1004 images and more comprehensive annotations than COCO, AMBER enables the detection of hallucinations beyond the 80 COCO categories, offering a broader evaluation scope.

\subsection{Prompt Design for EEScore}
\label{sec:appendix_ee_prompt}

For hallucination detection, we employ a ``reason-then-imagine'' prompt to derive both the imagination and reasoning sets used in the computation of EEScore (Sec.~\ref{sec:external}). The specific prompt are presented as follows:

{
\footnotesize
\begin{tcolorbox}
    \vspace{-1mm}
    Based on this image, please imagine what object might be in the \{direction\} outside the frame, and explain why. Specifically, your response should follow the following format: \\ \\
    Imagination: \textless one imaginary object here\textgreater \\
    Reason: The image features \textless briefly describe this image, be careful to mention all objects related to your imagination\textgreater, which suggests that \textless your imagination here\textgreater.
    \vspace{-1mm}
\end{tcolorbox}
}

\subsection{Construction and Insertion of Contrastive Contextual Tokens (CCT)}
\label{sec:appendix_cct}

After identifying the potential hallucinated objects $S_{induction}$ as described in the paper, we construct CCT by first truncating or padding the elements in this set to a fixed length $N$, yielding a new set $S'$, and then encoding them using a text encoder.

Specifically, when $|S_{induction}|>N$, \ie the number of elements in the potential hallucinated objects set exceeds $N$, the set is truncated based on the priority of each element, with the lowest-priority elements being removed. The priority assignment is determined as follows:
\begin{itemize}
    \item If both elements are sourced from IG, the element exhibiting the higher similarity in attention score is assigned higher priority.
    \item If one element originates from IG and the other from EE, the element from IG is given precedence.
    \item If both elements are sourced from EE, they are deemed to have equal priority, and removal is determined by a random selection process.
\end{itemize}

On the other hand, when $|S_{induction}| < N$, we randomly select additional \textit{unrelated} objects from a predefined object list to include in the set. Objects that have never appeared in our pipeline before, including the caption and EE responses, are considered unrelated.

To derive the CCT from \( S' \), we first concatenate all elements of \( S' \) into a single string using a predefined separator \( T_{sep} \). This ensures a structured and well-defined representation for encoding:
\[
    T = s_1 \ T_{sep} \ s_2 \ T_{sep} \ \dots \ T_{sep} \ s_N, \quad \text{where } s_i \in S'.
\]

Finally, we apply the text encoder \( \phi \) to generate the corresponding text embedding for the modified set \( S' \), which can be formally expressed as:
\[
    x_{cct} = \phi(T).
\]

For the insertion of the CCT, we place it in the contrastive decoding branch immediately after image tokens.

\subsection{Hyperparameters for Induction and Suppression}
\label{sec:appendix_hyperparameter}

\noindent \textbf{Hyperparameters for Induction.}
We consistently use greedy decoding when generating hallucination candidates. For the EE metric, we employed $|\mathcal{D}|=8$. The directions are: ``\textit{top}'', ``\textit{bottom}'', ``\textit{left side}'', ``\textit{right side}'', ``\textit{top left corner}'', ``\textit{top right corner}'', ``\textit{bottom left corner}'', and ``\textit{bottom right corner}''.

\noindent \textbf{Hyperparameters for Suppression.}
Across all experiments, the model is prompted with the instruction: ``\textit{Please help me describe the image in detail.}'' to generate captions. For nucleus sampling, we set the temperature to $1.0$ and $\text{top\_p}$ to $1.0$. In beam search, we used a beam size of 5. We employed nucleus sampling when evaluating AMBER. For all suppression experiments, we adapt different hyperparameters for different models (See Table~\ref{tab:parameters}.)

\section{Supplementary Experiments for Suppression}
\label{sec:appendix_exp}

Unless otherwise specified, all experimental results in this chapter are based on the LLaVA v1.5 7B model.

\begin{table}[t!]
    \centering
    \resizebox{1.\linewidth}{!}{%
        \begin{tabular}{llcccccc}
            \toprule
            EE         & IG         & $\mathrm{CHAIR}_S$$\downarrow$ & $\mathrm{CHAIR}_I$$\downarrow$ & Precision & Recall & F1     & Len     \\
                    \midrule
                       &            & $58.6$                         & $18.8$                         & $68.1$    & $76.4$ & $72.0$ & $105.2$ \\
            \checkmark &            & $51.0$                         & $14.4$                         & $73.9$    & $77.1$ & $75.5$ & $102.4$ \\
                       & \checkmark & $50.4$                         & $14.9$                         & $74.7$    & $76.6$ & $75.6$ & $100.3$ \\
            \checkmark & \checkmark & $48.6$                         & $14.5$                         & $74.6$    & $77.7$ & $76.1$ & $100.9$ \\
            \bottomrule
        \end{tabular}
    }
    \caption{Ablation study on CHAIR with LLaVA v1.5 7B}
    \label{tab:chair_abla}
\end{table}

\begin{table}[t]
    \centering
    \resizebox{1.\linewidth}{!}{%
        \begin{tabular}{clcccc|c}
            \hline
            \textbf{Dataset}         & \textbf{Setting}                      & \textbf{+ours} & Acc.$\uparrow$  & Prec.           & Recall          & F1$\uparrow$    \\ \hline
            \multirow{6}{*}{MSCOCO}  & \multirow{2}{*}{\textit{Random}}      & \ding{55}      & $85.0$          & $97.5$          & $71.8$          & $82.7$          \\
                                     &                                       & \Checkmark     & $\mathbf{86.3}$ & $\mathbf{98.7}$ & $\mathbf{73.6}$ & $\mathbf{84.3}$ \\ \cline{2-7}
                                     & \multirow{2}{*}{\textit{Popular}}     & \ding{55}      & $81.7$          & $89.5$          & $71.9$          & $79.7$          \\
                                     &                                       & \Checkmark     & $\mathbf{83.3}$ & $\mathbf{91.4}$ & $\mathbf{73.4}$ & $\mathbf{81.4}$ \\ \cline{2-7}
                                     & \multirow{2}{*}{\textit{Adversarial}} & \ding{55}      & $80.5$          & $86.8$          & $72.1$          & $78.7$          \\
                                     &                                       & \Checkmark     & $\mathbf{81.5}$ & $\mathbf{87.6}$ & $\mathbf{73.4}$ & $\mathbf{79.9}$ \\ \hline
            \multirow{6}{*}{A-OKVQA} & \multirow{2}{*}{\textit{Random}}      & \ding{55}      & $78.8$          & $96.3$          & $59.9$          & $73.9$          \\
                                     &                                       & \Checkmark     & $\mathbf{79.4}$ & $\mathbf{97.1}$ & $\mathbf{60.6}$ & $\mathbf{74.6}$ \\ \cline{2-7}
                                     & \multirow{2}{*}{\textit{Popular}}     & \ding{55}      & $76.1$          & $88.5$          & $60.0$          & $71.5$          \\
                                     &                                       & \Checkmark     & $\mathbf{76.9}$ & $\mathbf{89.5}$ & $\mathbf{61.0}$ & $\mathbf{72.6}$ \\ \cline{2-7}
                                     & \multirow{2}{*}{\textit{Adversarial}} & \ding{55}      & $72.5$          & $80.2$          & $59.9$          & $68.5$          \\
                                     &                                       & \Checkmark     & $\mathbf{73.9}$ & $\mathbf{82.7}$ & $\mathbf{60.5}$ & $\mathbf{69.9}$ \\ \hline
            \multirow{6}{*}{GQA}     & \multirow{2}{*}{\textit{Random}}      & \ding{55}      & $75.5$          & $94.1$          & $54.4$          & $58.9$          \\
                                     &                                       & \Checkmark     & $\mathbf{76.3}$ & $\mathbf{95.0}$ & $\mathbf{55.5}$ & $\mathbf{70.0}$ \\ \cline{2-7}
                                     & \multirow{2}{*}{\textit{Popular}}     & \ding{55}      & $71.2$          & $82.0$          & $54.3$          & $65.3$          \\
                                     &                                       & \Checkmark     & $\mathbf{71.7}$ & $\mathbf{82.1}$ & $\mathbf{55.5}$ & $\mathbf{66.2}$ \\ \cline{2-7}
                                     & \multirow{2}{*}{\textit{Adversarial}} & \ding{55}      & $69.6$          & $78.1$          & $54.5$          & $64.2$          \\
                                     &                                       & \Checkmark     & $\mathbf{70.2}$ & $\mathbf{78.6}$ & $\mathbf{55.5}$ & $\mathbf{65.1}$ \\ \hline
        \end{tabular}
    }
    \caption{%
        Results on POPE with LLaVA v1.5 7B. Acc. stands for accuracy, and prec. stands for precision. Higher scores indicate better performance and fewer hallucinations.
    }
    \label{tab:pope}
\end{table}

\subsection{Ablation Study}
\label{sec:appendix_abla}

In Table~\ref{tab:chair_abla}, we conduct an ablation study on the CHAIR benchmark to assess the contributions of different components in HalTrapper, namely External Expansion (EE) and Internal Grounding (IG). The baseline model without EE or IG achieves a CHAIR\(_S\) score of 58.6\% and a CHAIR\(_I\) score of 18.8\%. When adding EE alone, CHAIR\(_S\) reduces significantly to 51.0\%, while CHAIR\(_I\) decreases to 14.4\%. Precision improves to 73.9\%, Recall to 77.1\%, and F1 to 75.5\%, indicating a clear enhancement in reducing hallucinations and improving response quality. Incorporating IG alongside EE further decreases CHAIR\(_S\) to 50.4\% and slightly raises CHAIR\(_I\) to 14.9\%, showing that IG helps maintain high response quality with moderate gains in hallucination reduction. Finally, using both EE and IG achieves the best results, with CHAIR\(_S\) and CHAIR\(_I\) reduced to 48.6\% and 14.5\%, respectively. These findings confirm that the combination of EE and IG maximizes performance by effectively balancing precision, recall, and hallucination reduction, achieving the highest overall reliability and accuracy in the responses.

\begin{table}[t!]
    \centering
    \small
    \setlength\tabcolsep{5mm}
    \renewcommand\arraystretch{1}
    \resizebox{.9\linewidth}{!}{
        \begin{tabular}{lcc}
            \toprule
            \textbf{MM-Vet gen. subset} & \textbf{Baseline} & \textbf{Ours} \\
            \midrule
            LLaVA v1.5 7B               & 23.2              & 25.5          \\
            Qwen VL Chat                & 30.7              & 31.1          \\
            \bottomrule
        \end{tabular}
    }
    \caption{Results on MM-Vet~\cite{yu2024mmvetevaluatinglargemultimodal} generation subset.}
    \label{tab:mmvet}
\end{table}

\label{sec:appendix_mmvet}

\subsection{Additional Experiments on Adapted POPE}
\label{sec:appendix_pope}

POPE~\cite{Li-hallucination-2023}, the Polling-based Object Probing Evaluation (POPE) is aimed at evaluating hallucinations in LVLMs. In a manner similar to the CHAIR benchmark, POPE addresses object hallucinations by querying the model with prompt ``Is there a/an \{object\} in the image?'' to determine whether the model can correctly identify specific objects within images. The full POPE evaluation consists of three distinct subsets: the ``random'' subset, which tests objects randomly chosen from the dataset; the ``popular'' subset, which focuses on commonly occurring objects; and the ``adversarial'' subset, which challenges the model's ability to identify objects that are closely related to those actually present in the image.

Different from the general POPE evaluation pipeline, since our method is specifically designed for hallucinations in the context of long text, we adapted it's pipeline by reframing it as an image captioning task. Specifically, we first prompt the model to generate a detailed caption for each image and subsequently use the GPT-4o-mini model to assess whether the specified queried object appears in the caption. We have retained POPE's original evaluation metrics, such as recall and F1 score.

\begin{table*}[t]
\footnotesize
\centering
\begin{tabular}{p{0.95\linewidth}}
    \toprule
    GPT-4o Prompt
    \\
    \midrule
    You are required to score the performance of three AI assistants in describing a given image. You should pay extra attention to the hallucination, which refers to the part of descriptions that are inconsistent with the image content, such as claiming the existence of something not present in the image or describing incorrectly in terms of the counts, positions, or colors of objects in the image. Please rate the responses of the assistants on a scale of 1 to 10, where a higher score indicates better performance, according to the following criteria: \\
    1: Accuracy: whether the response is accurate with respect to the image content. Responses with fewer hallucinations should be given higher scores. \\
    2: Detailedness: whether the response is rich in necessary details. Note that hallucinated descriptions should not count as necessary details. \\
    3: Fluency: whether the response sound natural and well-phrased. Responses that avoid excessive repetition and awkward phrasing should receive higher scores. \\
    Please output the scores for each criterion, containing only three values indicating the scores for Assistant 1, 2 and 3, respectively. The three scores are separated by a space. Following the scores, please provide an explanation of your evaluation, avoiding any potential bias and ensuring that the order in which the responses were presented does not affect your judgment. \\ \\

    \text{[Assistant 1]} \\
    \{\} \\
    \text{[End of Assistant 1]} \\ \\

    \text{[Assistant 2]} \\
    \{\} \\
    \text{[End of Assistant 2]} \\ \\

    \text{[Assistant 3]} \\
    \{\} \\
    \text{[End of Assistant 3]} \\ \\

    Output format: \\
    Accuracy: \textless Scores of the three answers\textgreater \\
    Reason: \\ \\

    Detailedness: \textless Scores of the three answers\textgreater \\
    Reason: \\ \\

    Fluency: \textless Scores of the three answers\textgreater \\
    Reason: \\
    \bottomrule
\end{tabular}
\caption{The prompt used for GPT-4o evaluation.}
\label{tab:gpt4o_prompt}
\end{table*}

\noindent \textbf{Results.}
The results in Table~\ref{tab:pope} demonstrate that HalTrapper consistently enhances performance across all settings and datasets. For instance, on the MSCOCO~\cite{lin2014microsoft} dataset, HalTrapper achieves up to a 1.7\% improvement in F1 score in the ``popular'' setting, increasing from 79.7\% to 81.4\%. Similarly, on the A-OKVQA~\cite{schwenk2022okvqa} dataset, the model shows a gain of 1.4\% in the ``adversarial'' setting (from 68.5\% to 69.9\%). On the GQA~\cite{hudson2019gqa} dataset, the method delivers substantial improvements, with the F1 score increasing by 1.3\% in the ``popular'' setting (from 65.3\% to 66.2\%). These consistent gains highlight the effectiveness of HalTrapper in addressing hallucinations across various object recognition scenarios.

\subsection{Additional Experiments on MM-Vet}
MM-Vet~\cite{yu2024mmvetevaluatinglargemultimodal} is a benchmark designed to evaluate the response quality of LVLMs on complex multi-modal tasks. Questions in MM-Vet requires models to integrate multiple core capabilities. Given that our HalTrapper is designed for long response scenarios, we evaluate only the subset of MM-Vet questions that are explicitly annotated as assessing language generation and report the score. The evaluation is conducted using their official online evaluator.

\noindent \textbf{Results.}
Table~\ref{tab:mmvet} presents the performance of our HalTrapper compared to the baseline on the MM-Vet~\cite{yu2024mmvetevaluatinglargemultimodal} generation subset on both LLaVA v1.5 7B and Qwen VL. It can be observed that our HalTrapper achieves consistent improvements across two different models.

\subsection{Additional Results of GPT-4o Assisted Evaluation}
\label{sec:appendix_gpt}

Since the CHAIR metric only evaluates object-level hallucinations while ignoring other types, such as colors and numbers, following prior work~\cite{huang2024opera, liu2024paying}, we adapt GPT-4o~\cite{achiam2023gpt} for a more comprehensive assessment. GPT-4o's ability to perceive and interpret images allows it to evaluate hallucinations in longer responses, closely aligning with expert human judgment. Unlike previous studies that focused only on accuracy and detailedness, we expand the evaluation to include fluency, recognizing its importance in language generation. Specifically, we sample 50 images from COCO and prompt GPT-4o to score each generated text on a scale of 1-10. The exact prompt used is provided in Table~\ref{tab:gpt4o_prompt}.

\begin{table}[t!]
    \centering
    \small
    \setlength\tabcolsep{5mm}
    \renewcommand\arraystretch{1}
    \resizebox{.9\linewidth}{!}{
        \begin{tabular}{lccc}
            \toprule
            \textbf{GPT Eval} & \textbf{Baseline} & \textbf{PAI} & \textbf{Ours} \\
            \midrule
            Hal avg score     & 6.06              & 6.15         & 6.12          \\
            Det avg score     & 6.18              & 5.47         & 6.38          \\
            Flu avg score     & 7.56              & 7.38         & 7.59          \\
            \bottomrule
        \end{tabular}
    }
    \caption{Comparison between PAI~\cite{liu2024paying} and our HalTrapper on GPT-4o evaluation using the COCO~\cite{lin2014microsoft} dataset with LLaVA v1.5 7B.}
    \label{tab:gpt-pai}
\end{table}

\noindent \textbf{Results.}
Table~\ref{tab:gpt-pai} presents a comparison between our method and PAI~\cite{liu2024paying} in three evaluation dimensions using GPT: hallucination (Hal), detail (Det), and fluency (Flu). Our findings indicate that PAI currently leads in terms of reducing hallucinations and providing detailed responses. However, we observed that PAI often attempts to repeat content in order to influence GPT’s evaluation, leading to inflated Hal and Det scores that do not necessarily reflect genuine response quality. To address this, we introduced an additional Flu score to more comprehensively assess response quality and hallucination levels alongside Hal and Det scores. Our method achieves significantly more detailed and coherent text responses while maintaining a hallucination level comparable to that of PAI.

\section{Qualitative Results for Suppression}
\label{sec:appendix_visualization}

\subsection{Comparison with PAI}

\begin{figure*}[p!]
    \centering
    \includegraphics[width=1.\linewidth]{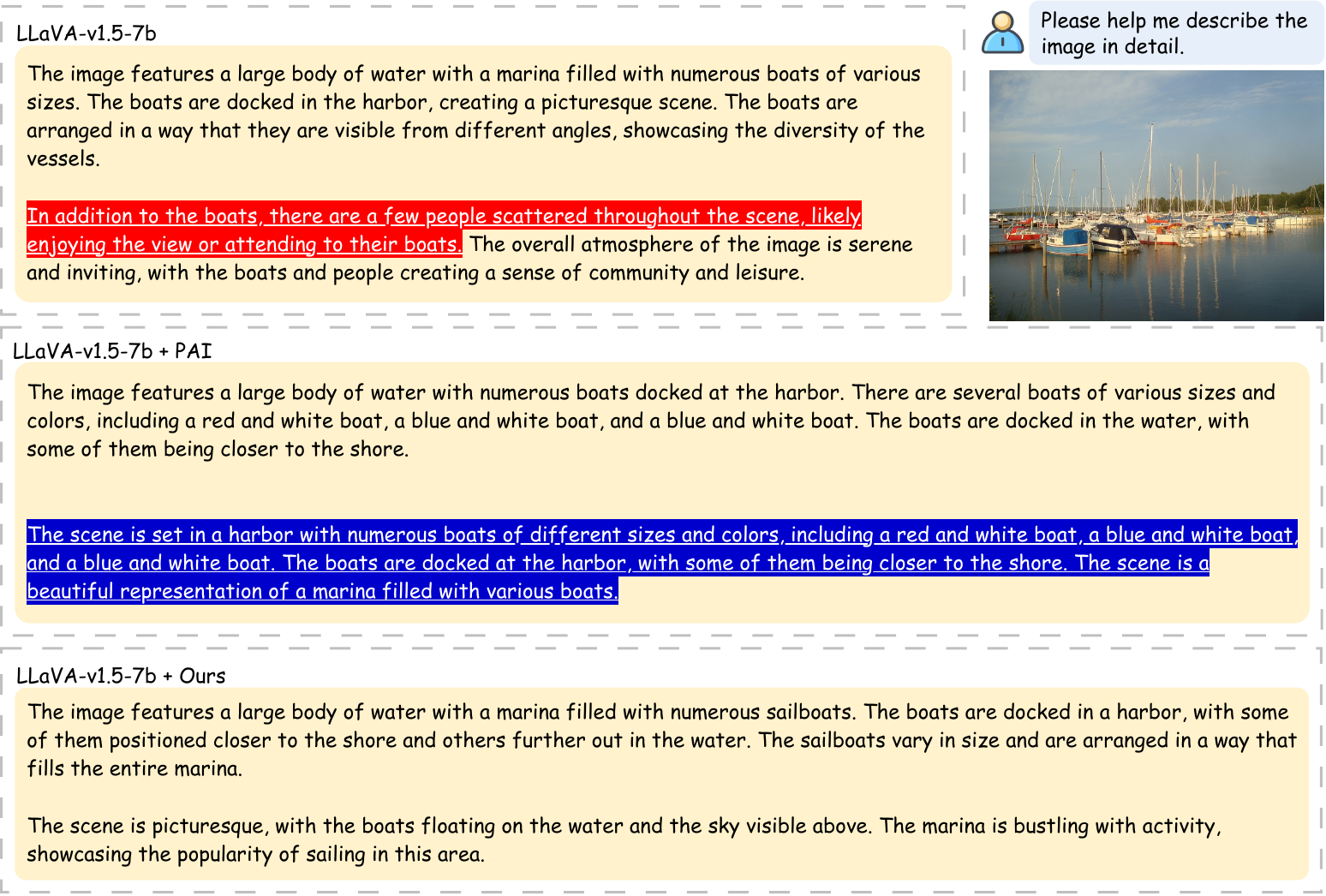}
    \caption{A qualitative comparison between baseline, PAI and our HalTrapper. Hallucinations are highlighted in \textcolor{red}{\textbf{red}} and the repetitions are highlighted in \textcolor{blue}{\textbf{blue}}.}
    \label{fig:compare_pai}
\end{figure*}

Although PAI~\cite{liu2024paying} demonstrates superior performance on hallucination benchmarks, its approach of directly enhancing attention scores adversely affects the model's language generation capabilities. Specifically, after applying the PAI method, LVLMs tend to produce redundant information. This issue is illustrated in Table~\ref{tab:gpt-pai}, which presents evaluations using GPT-4o. We also present illustrative examples provided in Fig.~\ref{fig:compare_pai}.

We observe that PAI poses a risk of redundantly repeating image content when generating descriptions. For instance, details such as ``boats docked at the harbor," ``a red and white boat, a blue and white boat, and a blue and white ship," and ``some boats are closer to the shore" are frequently reiterated across consecutive sentences. This redundancy compromises the coherence and logical structure of the generated output. In contrast, our model effectively mitigates such hallucinations, such as “a few people”, while maintaining both the logical consistency and content integrity of the description.

\subsection{Qualitative Results of Our HalTrapper}

\begin{figure*}[p!]
    \centering
    \includegraphics[width=1.\linewidth]{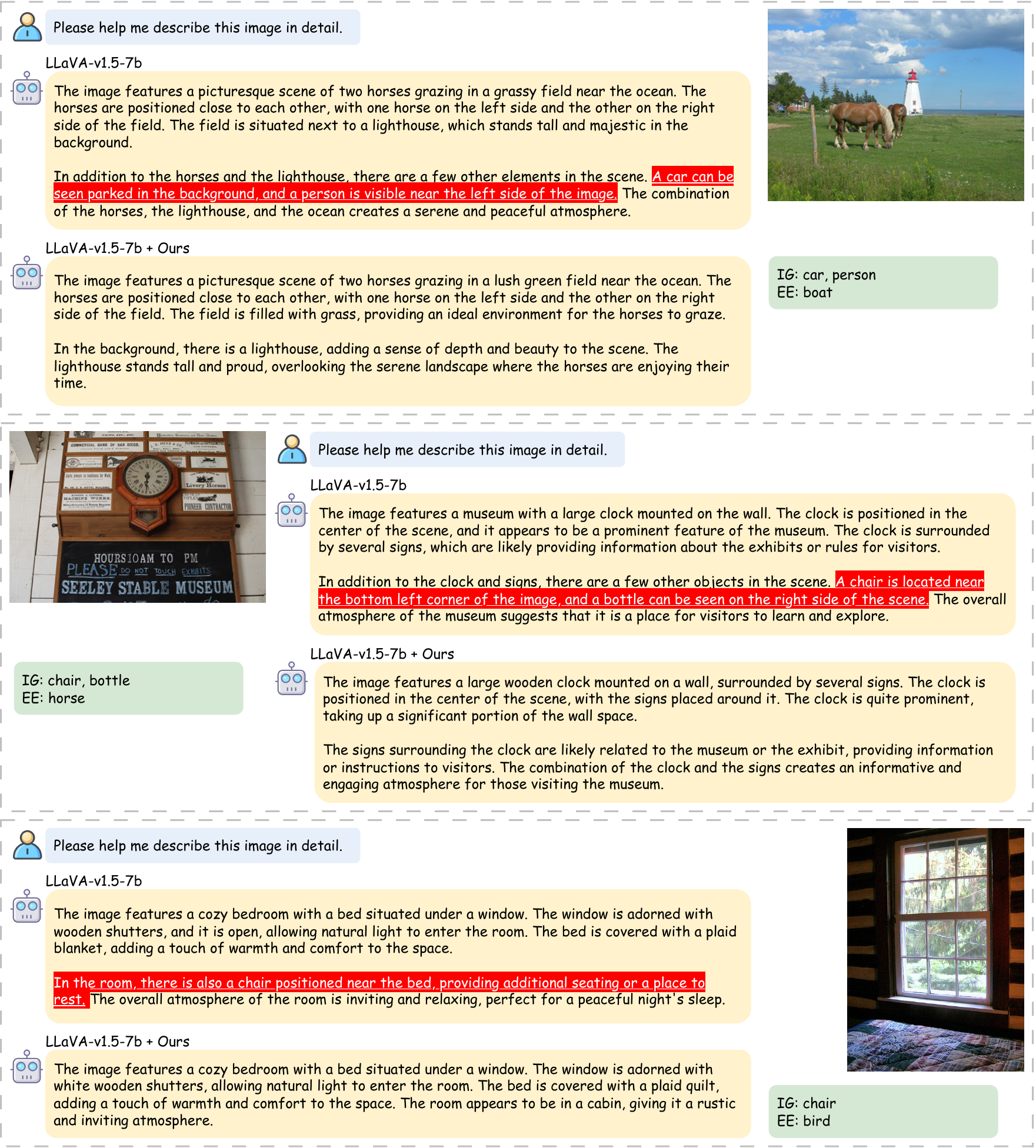}
    \caption{Examples of hallucination suppression using our HalTrapper. Hallucinations are highlighted in \textcolor{red}{\textbf{red}}.}
    \label{fig:visual1}
\end{figure*}

\begin{figure*}[p!]
    \centering
    \includegraphics[width=1.\linewidth]{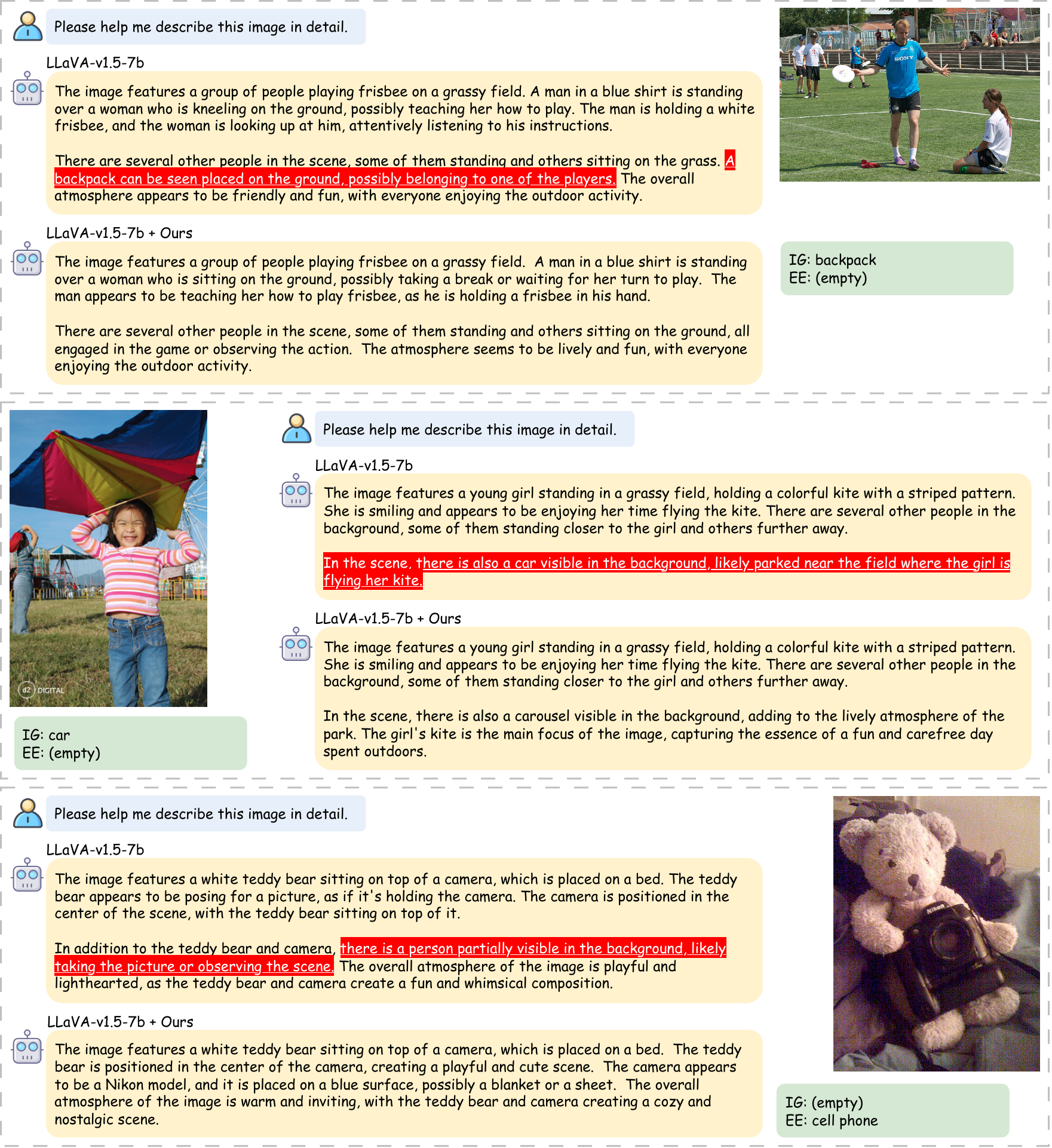}
    \caption{Examples of hallucination suppression using our HalTrapper. Hallucinations are highlighted in \textcolor{red}{\textbf{red}}.}
    \label{fig:visual2}
\end{figure*}

We provide additional visualizations to further demonstrate the effectiveness of our method, as shown in Fig.~\ref{fig:visual1} and~\ref{fig:visual2}.

These results highlight the effectiveness of our proposed method. Specifically, the hallucinated objects generated by IG exhibit a notable overlap with the ground truth hallucinations in the caption, while our Contrastive Contextual Decoding (CCD) process effectively mitigates these hallucinations. In contrast, considering the issue of false positives, EE avoids the direct incorporation of hallucinated objects in captions. However, it still contributes to hallucination suppression. As demonstrated in the final example of Fig.~\ref{fig:visual2}, even though EE does not directly include the object ``person," it extracts a latent, hallucinated object ``cell phone," which is closely related to ``person," thereby preventing the model from hallucinating ``person."

\section{Limitations}
\label{sec:appendix_limitations}

This work primarily addresses object-level hallucinations in long-form responses generated by large LVLMs. However, LVLMs are susceptible to a broader spectrum of hallucinations, including failures in instruction following and hallucinations at the attribute and relational levels. Moreover, our evaluations are mainly on image captioning benchmarks such as CHAIR and AMBER. While these benchmarks are widely used for evaluating hallucinations, they do not adequately cover more open-ended generative scenarios. Developing more comprehensive and standardized benchmarks for such settings represents a valuable direction for future research.

\clearpage
\newpage

\end{document}